\pdfoutput=1
\documentclass[letterpaper, 10 pt, conference]{ieeeconf}  

\IEEEoverridecommandlockouts                              

\usepackage{lmodern}
\usepackage{amsmath} 
\usepackage{amsfonts}  

\usepackage{graphics} 
\usepackage{graphicx} 
\usepackage{epsfig} 
\usepackage{bm} 
\usepackage{times} 
\DeclareMathOperator*{\argmin}{arg\,min}  
\usepackage{amssymb}  
\usepackage{bm}
\usepackage{algorithm}
\usepackage{algpseudocode}
\usepackage{mathtools}
\usepackage{listings}
\usepackage{subcaption}
\usepackage{multirow}
\usepackage{hyperref}
\usepackage[table,xcdraw]{xcolor}
\usepackage{optidef}
\usepackage{gensymb}

\newcommand{\br}[1]{\mathbf{\bm{#1}}}

\usepackage{hyperref}

\begin{document}
\title{\LARGE \bf
Toward Control of Wheeled Humanoid Robots with Unknown Payloads: Equilibrium Point Estimation via Real-to-Sim Adaptation
}

\author{Donghoon Baek$^{1}$, Youngwoo Sim$^{1}$, Amartya Purushottam$^{2}$, Saurabh Gupta$^{2,3}$, and Joao Ramos$^{1,2}$
\thanks{This work is supported by University of Illinois, 1-200250-917015-917396.}
\thanks{The authors are with the $^1$ Department of Mechanical Science and Engineering and the $^2$ Department of Electrical and Computer Engineering and the $^3$ Department of Computer Science at the University of Illinois at Urbana-Champaign, USA.{\tt\small dbaek4@illinois.edu}} 
}

\maketitle

\begin{abstract}
Model-based controllers using a linearized model around the system's equilibrium point is a common approach in the control of a wheeled humanoid due to their less computational load and ease of stability analysis. However, controlling a wheeled humanoid robot while it lifts an unknown object presents significant challenges, primarily due to the lack of knowledge in object dynamics. This paper presents a framework designed for predicting the new equilibrium point explicitly to control a wheeled-legged robot with unknown dynamics. We estimated the total mass and center of mass of the system from its response to initially unknown dynamics, then calculated the new equilibrium point accordingly. To avoid using additional sensors (e.g., force torque sensor) and reduce the effort of obtaining expensive real data, a data-driven approach is utilized with a novel real-to-sim adaptation. A more accurate nonlinear dynamics model, offering a closer representation of real-world physics, is injected into a rigid-body simulation for real-to-sim adaptation. The nonlinear dynamics model parameters were optimized using Particle Swarm Optimization. The efficacy of this framework was validated on a physical wheeled inverted pendulum, a simplified model of a wheeled-legged robot. The experimental results indicate that employing a more precise analytical model with optimized parameters significantly reduces the gap between simulation and reality, thus improving the efficiency of a model-based controller in controlling a wheeled robot with unknown dynamics.

\end{abstract}

\section{Introduction}
\label{S:1}
Wheel-legged robots have emerged as a potential platform to facilitate operations in a wide range of industries, including agriculture, mining, military, and search and rescue. Combining the mobility of wheels and the mobility of legs, the capability to navigate difficult environments could be greatly improved. \cite{klemm2020lqr, purushottam2023dynamic, purushottam2022hands, BostonDynamics, baek2023study}.

Controlling a wheeled-legged robot poses significant challenges due to its inherent instability and complex nonlinear dynamics. Traditional methods often simplify the system by linearizing the system dynamics around its equilibrium point, enabling the use of diverse control techniques like the Linear Quadratic Regulator (LQR) or Model Predictive Control (MPC) \cite{klemm2020lqr,bjelonic2021whole}. The underlying assumption is that the system can be treated as a linear system within the equilibrium point. For example, within the equilibrium point, the inverted pendulum remains upright and stationary, meaning that the net force acting on the pendulum is zero and it does not experience any acceleration. However, when the robot engages in tasks like manipulating objects or pushing heavy loads, these changes can drastically alter the equilibrium point due to shifts in the center of mass (CoM) or contact forces, making linear approximations inadequate for effective control \cite{sonnleitner2019mechanics}. Recently, approaches such as deep reinforcement learning (RL) and domain randomization have emerged, showing a promising result in locomotion tasks \cite{baek2022hybrid,cui2021learning,hwangbo2019learning}. Yet, they suffer from interpretability issues and cannot assure stability.

\begin{figure}[t!]
\begin{center}
\includegraphics[width=1\linewidth]{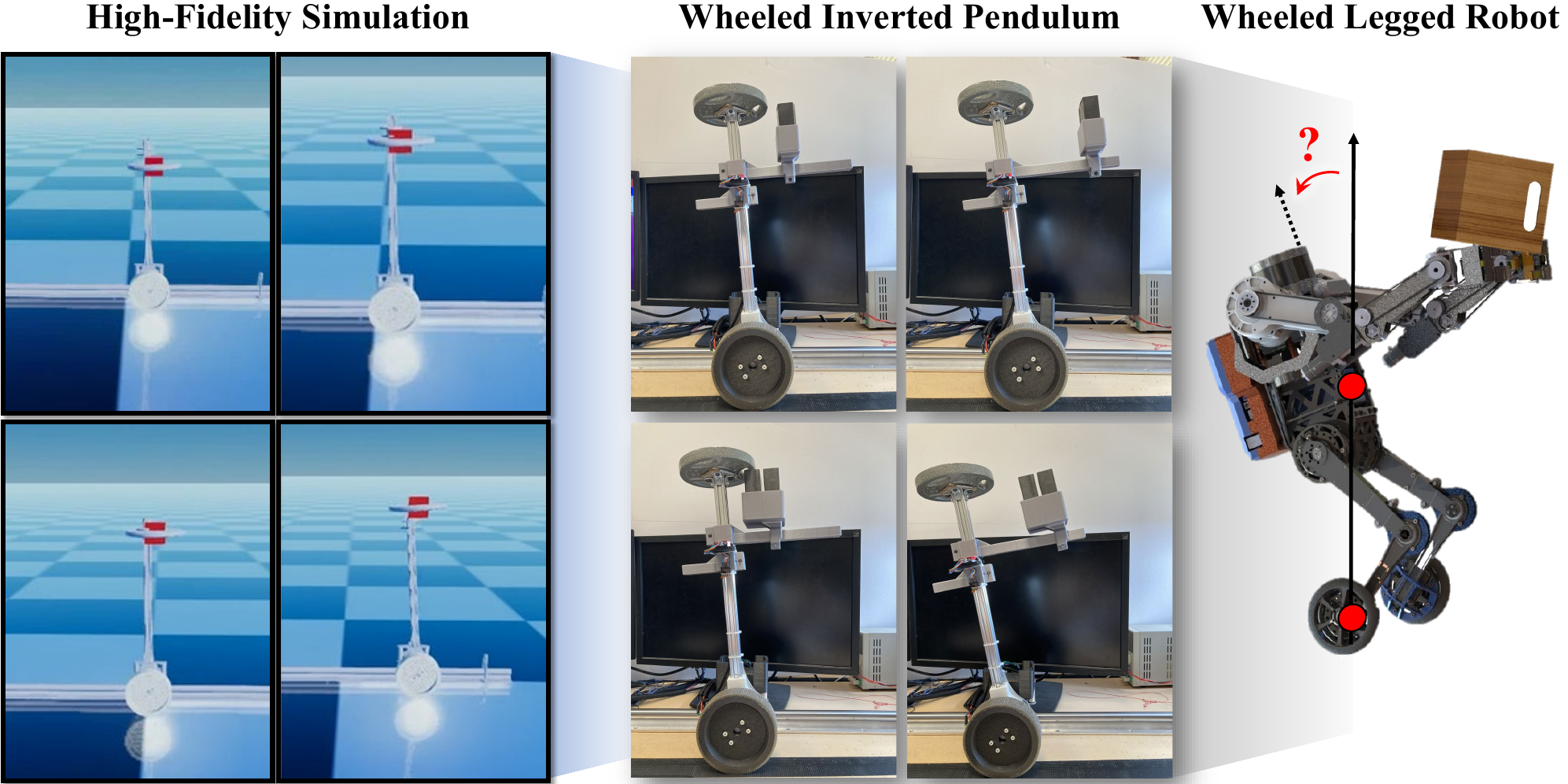}
\end{center}
\caption{\textbf{Conceptual overview of the proposed method.} The data-driven regression model learns to identify the new equilibrium point for a wheeled inverted pendulum from a high-fidelity simulation closely mirroring the real world. A physical version of this pendulum tests the framework's viability, suggesting its applicability in controlling a wheeled humanoid robot lifting an unknown object.}
\label{fig1_concept}
\vspace{-1.5em}
\end{figure}

In this work, our focus is on addressing the challenge of a wheeled humanoid robot transporting an unseen object during locomotion. The framework capable of rapidly and explicitly estimating the new equilibrium point of a wheeled-legged robot is proposed. The main idea is to directly identify the equilibrium point by capturing the momentary change in a history of initial proprioceptive states when the robot falls (e.g., the heavier and farther away the object, the faster the robot will fall). Our approach does not involve solving the Newton-Euler equations, which are typically employed in system identification processes. As a result, we avoid the need for using force/torque sensors and cameras to estimate the new center of mass and total mass when an object is included. Moreover, our framework can simply be incorporated with a classical model-based controller without re-formulating their equation. 

A prominent challenge in adopting data-driven approaches is the issue of data scarcity. To counter this, we built a high-fidelity simulation via real-to-sim adaptation and trained a data-driven model in the simulation. The adaptation method includes a more accurate dynamics model that accounts for friction, damping, and actuator dynamics, elements that are often oversimplified in typical rigid-body dynamics simulations.

\section{Related Work}
\label{sec:citations}

\noindent \textbf{Wheeled-Legged Robot Control:}
The most popular approach to control wheeled-legged robots is model-based control, which leverages a mathematical dynamics model. Some strategies use simplified models, like the Wheeled Inverted Pendulum (WIP), and linearize them around the system's equilibrium point \cite{purushottam2022hands,purushottam2023dynamic, baek2023study}. To achieve better tracking and versatility, more complex models have been applied \cite{klemm2020lqr,chen2020underactuated,xin2020online}. However, the success of these methods depends greatly on the precision of the modeled dynamics, a precision that is difficult to attain and may not adjust well to changes in the system or environment. As an alternative, adaptive controllers have been presented to enable the robots to adapt to new environments \cite{minniti2021model,li2023multi}. Moreover, learning-based methods, combining deep neural networks with reinforcement learning (RL), have been suggested as an effective approach to address the nonlinear locomotion challenges of legged robots \cite{cui2021learning, baek2022hybrid}. With the benefits of simulation for safe and efficient training, many RL studies employ sim-to-real techniques, resulting in successfully transferring the RL policy to the real world \cite{lee2020learning, kumar2022adapting, tan2018sim}. Despite their innovative success, they often require a lot of manual tuning or have many difficulties in interpreting the behavior and ensuring the safety of the system. \\

\noindent \textbf{Sim-to-Real Transfer:} Despite learning-based approaches making controller design in simulations simpler and less reliant on specialized domain knowledge, sim-to-real transfer still necessitates extensive manual adjustments. For example, the process of choosing dynamic parameters to randomize for training a robust RL policy across diverse dynamics remains complex and not easily automated \cite{kumar2022adapting, tan2018sim, peng2020learning}. An alternative way to reduce a \textit{reality gap} is system identification which entails adjusting model parameters to align the simulation's observations with those from actual hardware \cite{muratore2021data,allevato2020tunenet,chebotar2019closing,jiang2021simgan,ramos2019bayessim}. While this approach has helped generate robust policies, the focus lies solely on adapting the physics engine's parameters or distribution which are often inaccurate due to their simplification. Many rigid-body simulators prioritizing fast computation often employ simplified dynamics to lessen computational demands, typically overlooking factors like static and viscous friction at joints, damping, and actuator dynamics. While some studies have explored data-driven approaches to model actuator dynamics \cite{hwangbo2019learning, lee2020learning}, these are not universally applicable to passive joints or varied controller types, such as position or torque modes. Instead of merely adjusting existing simulator parameters, our approach integrates a more detailed dynamics model with its optimized parameters into the simulation, aiming to minimize the \textit{reality gap}.

\section{BACKGROUND}
\label{B:1}
The qualitative behavior of the nonlinear system in the vicinity of a hyperbolic equilibrium point is determined by the characteristics of the corresponding linear system \cite{perko2013differential}. Given that \(x_0 = 0\) is an equilibrium point of the system, we have \(f(0) = 0\). By applying Taylor's Theorem, we can express the function \(f(x)\)in the expanded form as follows:

\begin{equation}\label{tayler}
\dot{x} = f(x) = Df(0)x + \frac{1}{2}D^2f(0)(x, x) + \ldots
\end{equation}

\noindent where \(Df(0)x\) denotes the Jacobian matrix of \(f\) evaluated at \(0\), linearly acting upon \(x\), and \(D^2f(0)(x, x)\) indicates the higher-order terms. The same theory can be applied to a wheeled-humanoid robot (e.g., SATYRR \cite{purushottam2023dynamic}) to linearize its nonlinear dynamics. The wheeled inverted pendulum is commonly taken to express the motion of a wheeled humanoid and its equation of motion is as follows:
\begin{equation}
\label{wip_dyn}
    \begin{gathered}[b] 
        \bigg (m_{b} \!+\! m_{w} \!+\! \frac{I_w}{r^2}\bigg )\ddot{x}_w \!+\! m_{b}L s({\theta})\dot{\theta}^2 
    \!-\! m_{b}L c({\theta})\ddot{\theta} \!= \!u\\
        (m_{b}L^2 + I_{b})\ddot{\theta} - m_{b}Lc({\theta})\ddot{x}_w - m_{b} gLs({\theta}) \!= \!0 
    \end{gathered}\
\end{equation}

\noindent where $m_{b}, m_w, L, \theta, I_w, I_{b}, r,$ and $u$ denote the body and wheel masses, the distance between the CoM to the center of the wheel, pitch angle, wheel and body inertia, wheel radius, and control force, respectively. Linearizing the system's dynamics at the equilibrium (CoM above the wheel) yields a linear state-space model:
\begin{equation}
    \Delta \dot{\br{q}} = \br{A}\Delta \br{q} + \br{B}\Delta u,
\end{equation}
where $\br{q} = [x_w, \theta, \dot{x}_w, \dot{\theta}]^\top$ defines the state vector, and $\br{A} \in \mathbb{R}^{4 \times 4}$, $\br{B} \in \mathbb{R}^{4 \times 1}$ are state-space matrices. Deviations from equilibrium state $\br{q_0}$ and nominal control effort $u_0$ are $\Delta \br{q} = \br{q} - \br{q_0}$ and $\Delta u = u - u_0$, respectively. Changes in total mass and equilibrium point $\theta_0$, such as when lifting an object, affect these dynamics, degrading model-based controller performance. An LQR controller $\pi_H$ designed with the above dynamics serves as our baseline for evaluating the proposed framework.

\section{Method}
\label{method}

\begin{figure*}[t]
\centerline{\includegraphics[width=10cm]{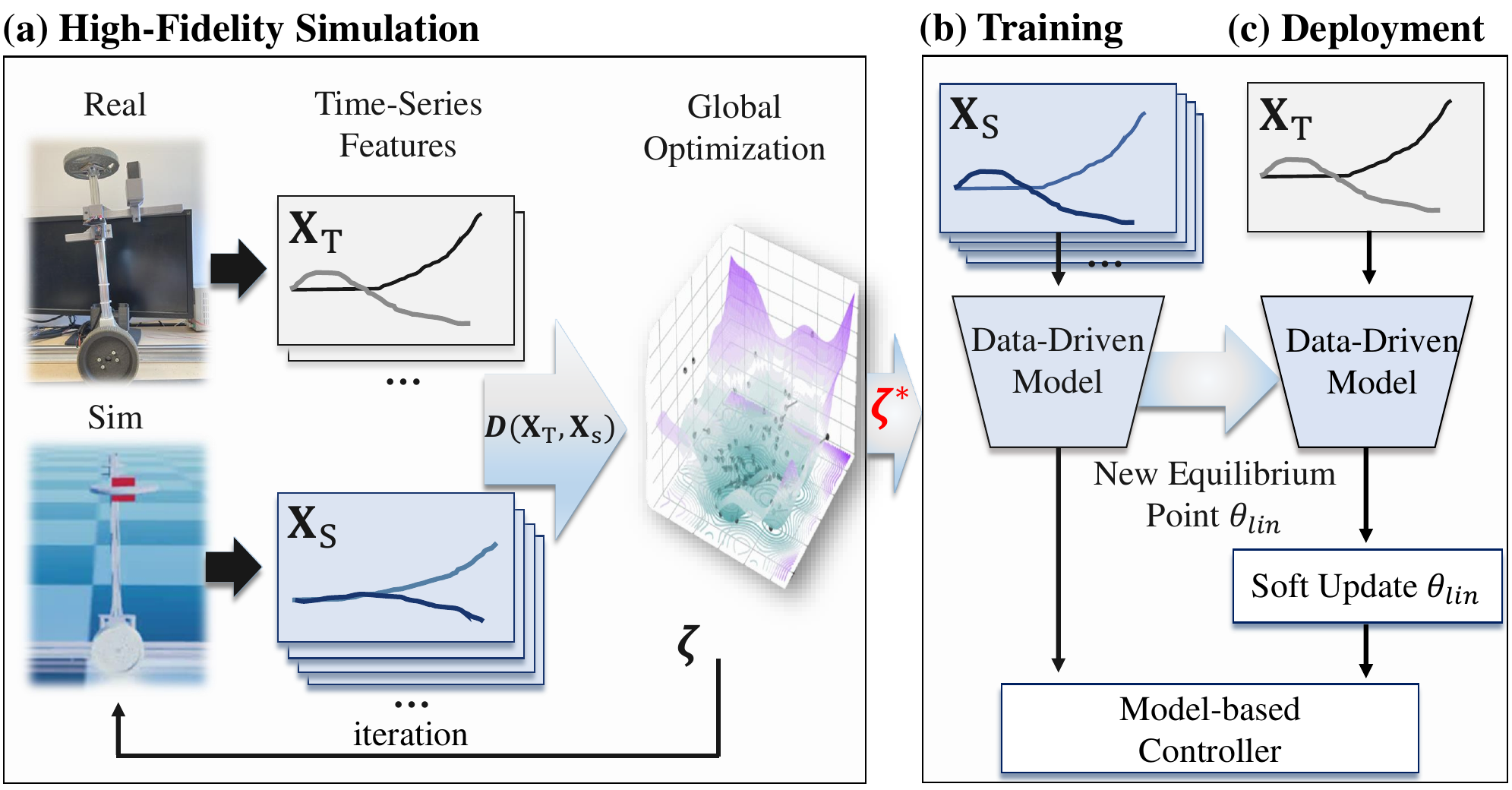}}
    \caption{\textbf{Real-to-Sim Adaptation via a High Fidelity Simulation.} (a) High-fidelity simulation is achieved by minimizing parametric modeling error, $\Delta \br{x_p}$, via updates to $\br{\zeta}$. The state trajectories of the system in two different domains are obtained from a physical system and a simulation, respectively, offline. The global optimization algorithm (e.g., PSO) is utilized to identify the parameters $\br{\zeta}$ to deduce the \textit{reality gap}. (b) A data-driven model (e.g., LSTM) is trained to predict the new equilibrium point via supervised learning with the data obtained from a high-fidelity simulation. (c) The trained data-driven model estimates the new equilibrium point online in deployment.}
    \label{fig2}
\vspace{-1.5em}
\end{figure*}

\subsection{Real-to-Sim Adaptation via a High-Fidelity Simulation}
Unlike previous studies that apply the domain transfer to a trained model to bridge the domain gap $\Delta \br{x}$ during or after training, our approach aims to reduce the \textit{reality gap} at the beginning of the entire process. This can be achieved by incorporating a more precise dynamics model that accounts for nonlinear friction, damping, and motor dead-zone effects into a rigid-body simulation. Especially, in this work, we are interested in minimizing the gap caused by the parametric modeling error $\Delta \br{x_p}$. 
\begin{equation}
    \Delta \br{x}(t) = \br{x}(t) - \hat{\br{x}}(t) = \underbrace{\Delta \br{x_p}(\br{\zeta}, t)}_{\text{Parametric Error}} + \underbrace{\Delta \br{x_{np}}(\br{x}, t)}_{\text{Nonparametric Error}}
\label{sim2real_gap_eqn}
\end{equation}
Here, $\Delta \br{x}(t)$ denotes the \textit{reality gap}, with $\Delta \br{x_p}$ as the parametric error dependent on model parameters $\br{\zeta}$ and time $t$, and $\Delta \br{x_{np}}$ representing nonparametric error related to state $\br{x}$ and time $t$. Particle Swarm Optimization (PSO)\cite{kennedy1995particle} is used for parameter optimization. A schematic of the framework is depicted in Fig. \ref{fig2}. \\

\label{sec:method1} 
\noindent \textbf{Data Construction.} To build a high-fidelity simulator toward achieving a real-to-sim adaptation, we constructed the $M$ number of source dataset $\mathcal{D_S} = { (\br{X_S}^i,y_S^i)}^{M}_{i=1}$, and a small $m$ number of target dataset $\mathcal{D_T} = { (\br{X_T}^i,y^i_T)}^{m}_{i=1}$ obtained from a simulator and a real-world, respectively ($m=40, M=1200$). Both domains acquire samples drawn from the source and target distribution, $\mathcal{D}^s \sim p_s(\br{X_S},y_S) $ and $\mathcal{D}^t \sim p_t(\br{X_T},y_T)$ which typically differ ($\mathcal{D}^s \in \mathcal{D_S}$ and $\mathcal{D}^t \in \mathcal{D_T}$). In our application, the output $y$ indicates the linearized pitch angle $\theta$ deciding the equilibrium point. Inspired by the previous research \cite{tan2018sim}, the input data $\br{X}$ consists of the state vector $\br{X} = [x_{t: T}, \theta_{t: T}]^\top$ represents a $T$ time-series trajectory of linear position and pitch angle. Since the goal is to estimate the new equilibrium point of a wheeled-legged robot in a situation where the total payload and CoM position change, we randomized the total mass, inertia, and CoM position of the robot while collecting the dataset $\mathcal{D}$. We utilized a RaiSim \cite{hwangbo2018per} simulation to construct the simulation dataset $\mathcal{D_S}$. Note that the new angle $\theta_{lin}$ is easily computed in a simulation with the known position of CoM. In the real world, we manually measured the $\theta_{lin}$ at which the robot can stabilize itself without controller intervention (see Fig. \ref{testbed}). To get more accurate and less noisy data in the real world, we applied the Extended Kalman Filter to a physical system and utilized the same control gain in both domains.\\

\begin{figure}[b]
\begin{center}
\includegraphics[width=0.9\linewidth]{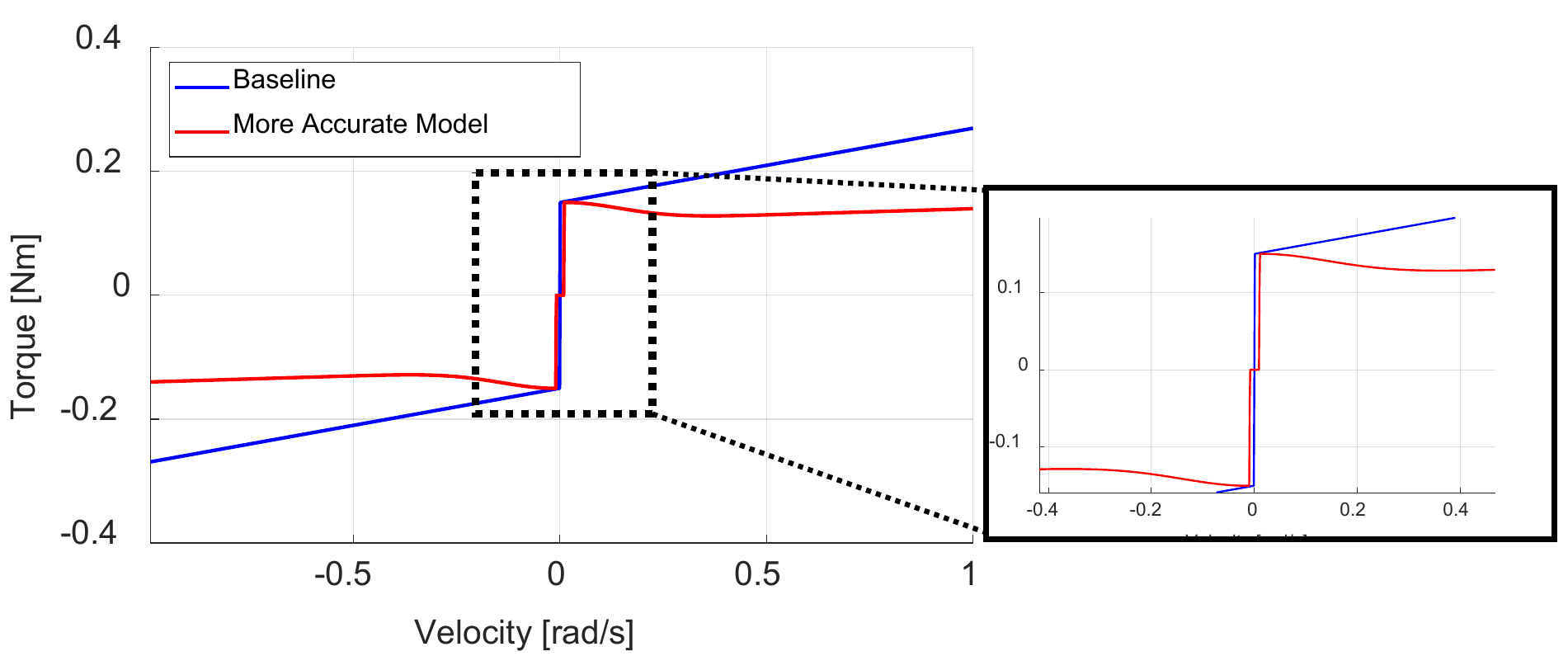}
\end{center}
\caption{\textbf{Visualizing the Nonlinear Dynamic Function $g_{\br{\zeta}}$.} The selected parameter $\br{\zeta}$ is [0.15, 0.12, 0.2, 0.02, 0.01, 0.7]. The model exhibits non-responsiveness near zero speed due to the dead zone effect, with a behavior that is more nonlinear compared to the standard viscous and Coulomb friction model (Baseline). Additionally, a slight phase shift is observed, indicative of a delay.}
\label{friction_model_fig}
\end{figure}

\noindent \textbf{More Accurate Nonlinear Dynamic Function.} The nonlinear dynamics model is designed by combining the core elements typically factored in during a sim-to-real transfer \cite{lee2020learning,kumar2022adapting,tan2018sim}. Since the success of most machine learning models (e.g., Gaussian Processes, Deep Neural Networks) highly depends on the quantity and quality of the dataset, we adopted to use of an analytical model with its optimized parameters considering the sample efficiency to reduce the \textit{reality gap}. Although some rigid-body simulators have a built-in function allowing users to tune the parameter of friction and damping, their function is usually simplified to reduce the complexity and sometimes not clear in deciding the numerical value. To use a more accurate and intuitive dynamic model in a simulation, we designed the nonlinear dynamics model that takes account of the effect of the static friction, damping, latency, and actuator dynamics in each joint. Here, a more accurate is denoted by $g_{\br{\zeta}}$ which satisfies $\br{x_{t+1}} = f(\br{x_t}, \br{u_t}, g_{\br{\zeta}}(\br{x_t}))$, where $\br{\zeta}=(F_{ss}, F_c, vs, \omega, \epsilon, \alpha)$ is physics parameters of the model $g_{\br{\zeta}}$. The nonlinear dynamics model $g_{\br{\zeta}}$ is designed as follows:
\begin{equation}
\begin{split}
    &g_{\br{\zeta}}(t) = -\alpha F_{ss}(t) + (\alpha-1)F_{ss}(t-1) \\
    &F_{ss} \!\! \, = \, \!\! 
    \begin{cases} 
      0 & \dot{q} < \epsilon \\
      F_{c} \text{sign}(v) + (F_{s} - F_{c}) e^{r} \text{sign}(v) + \sigma v & \dot{q} >= \epsilon
    \end{cases} \\
    &e^{r} = -(v/vs)^2
\label{nonlinear_f}
\end{split}
\end{equation}
\noindent where $\alpha$ represents the damping ratio, which delays signals, and $F_{ss}$ the static and dynamic friction model, incorporating a dead zone effect denoted by $\epsilon$. The friction model, $F_{ss}$, integrates Coulomb friction $F_c$, the differential static friction, and viscous damping $\sigma$, where the damping force correlates with velocity (see Fig. \ref{friction_model_fig}). The Stribeck velocity, $vs$, marks the shift from static to dynamic friction. Notably, the friction force, $F_{ss}$, is significant in steady states with minimal movement $v$,

\noindent leading to instability. To mitigate this, we introduce a deadzone $\epsilon$ in Eq. (\ref{nonlinear_f}). In our application, the model $g_{\br{\zeta}}$ is specifically applied to translation and actuator joints separately to address high friction and latency issues. Moreover, we applied the artificial high-frequency noise to acquire more realistic observed state $\br{X_S}$. The calculation for this high-frequency noise denoted as $w$, is presented subsequently.
\begin{equation}
    w = Acos(2\pi ft)\times B\mathcal{N}(0,1)
\label{noise}
\end{equation}
where $A, B$, and variables $f, t, \mathcal{N}(0,1)$ representing frequency, time, and a normally distributed random number, respectively (setting $A=0.01, B=0.5, f=0.0002$), noise $w$ is added to the original observation. We also chose an inertia value more carefully bu calculating the moment of inertia (e.g., $I = ML^2$) rather than using an arbitrarily selected value within a specific range. \\

\noindent \textbf{Parameter Optimization For Nonlinear Function via Particle Swarm Optimization.} 
If the accumulation of model match over trajectories $\br{X^t_S}$ and $\br{X^t_T}$ is the same, we can say that there is almost no \textit{reality gap} in two domains. Therefore, the objective is to find the most appropriate parameters $\br{\zeta}$ resulting in matching two trajectories as close as possible. We leverage mean square error (MSE) as a distance metric $d(\cdot|\cdot)$ across $m$ samples, each spanning $T$ units, to effectively bridge the \textit{reality gap} \cite{allevato2020tunenet}. 

\begin{equation}
    \argmin_{\zeta} \dfrac{1}{m}\sum_{i=1}^{m}\sum_{t=1}^{T}d(\br{X^t_S},\br{X^t_T}).
    \label{costfunc}
\end{equation}
The optimal parameter $\br{\zeta^*}$ can be chosen by solving the Eq. \ref{costfunc} and we utilized the PSO algorithm due to its global search ability and fast convergence speed. Real-to-Sim adaptation procedure is described in Algorithm \ref{alg:cap}.

    
    
    
\begin{algorithm}[t]
    \caption{Real-to-Sim Adaptation Procedure}\label{alg:cap}
    \begin{algorithmic}
    \State \textbf{Input:} $\mathcal{D_T} = \{D^1_T, D^2_T, \dots , D^m_T\}, m = 8$
    \State \textbf{Initialize:} $\br{\zeta} \in\mathbb{R}^k$, $\mathcal{L}\in \mathbb{R}, k=12$ \Comment{Initialize parameter of $g_{\br{\zeta}}$ and Loss function}
    \While{until $ (\mathcal{L} \text{  converges})$}  
        \State $({\br{X_T}^i,y_T^i)}^{m}_{i=1} \gets \mathcal{D_T} $ \Comment{From a real-world}
        \For{$i=0,...m$} \Comment{Run Raisim Simulator}
            \State $({\br{X}^i_S,y^i_S)}^{m}_{i=1} \gets \pi_H(\br{\zeta}) $ \Comment{From a baseline LQR }
        \EndFor
        \State Calculate the Loss function $\mathcal{L}$ \Comment{Eq. (\ref{costfunc})}
        \State Update parameter $\br{\zeta}$ via \textit{PSO algorithm} 
    \EndWhile
    \State \textbf{Return:} $\br{\zeta^*}$
    \end{algorithmic}
\end{algorithm}

The PSO algorithm, implemented via the SciPy package \cite{virtanen2020scipy}, updates particle velocities and positions as per:
\begin{equation}
\begin{split}
&V_{id}^{t+1} = wV_{id}^{t} + c_1r_1(P_{b, id}^{t} - X_{id}^{t}) + c_2r_2(G_{b, id}^{t} - X_{id}^{t}), \\
&X_{id}^{t+1} = X_{id}^{t} + V_{id}^{t+1}
\end{split}
\label{pso}
\end{equation}
Here, $V_{id}^{t+1}$ and $X_{id}^{t+1}$ denote the $i$th particle's velocity and position in dimension $d$ at time $t+1$. Parameters include inertia weight $w$, and cognitive and social factors $c_1$ and $c_2$, influencing the particle's momentum and the pull towards its best and the swarm's best positions, respectively, with $P_{b, id}^{t}$ and $G_{b, id}^{t}$ representing these best positions in dimension $d$ at time $t$ ($c1=0.5, c2=0.2, w=0.9$).

In the optimization of parameter \(\br{\zeta}\), the mean \(\br{X^{i}_T} \in \mathbb{R}^{N \times L}\) of real data per class is utilized, where \(N=2\) (features) and \(L=80\) (window size). The parameter range for \(\br{\zeta}\) in global optimization is manually tuned. For optimizing the built-in simulation parameters, seven parameters including joint friction and damping, crucial for \textit{domain randomization}, are selected.



\begin{figure*}[t]
\centerline{\includegraphics[width=15cm]{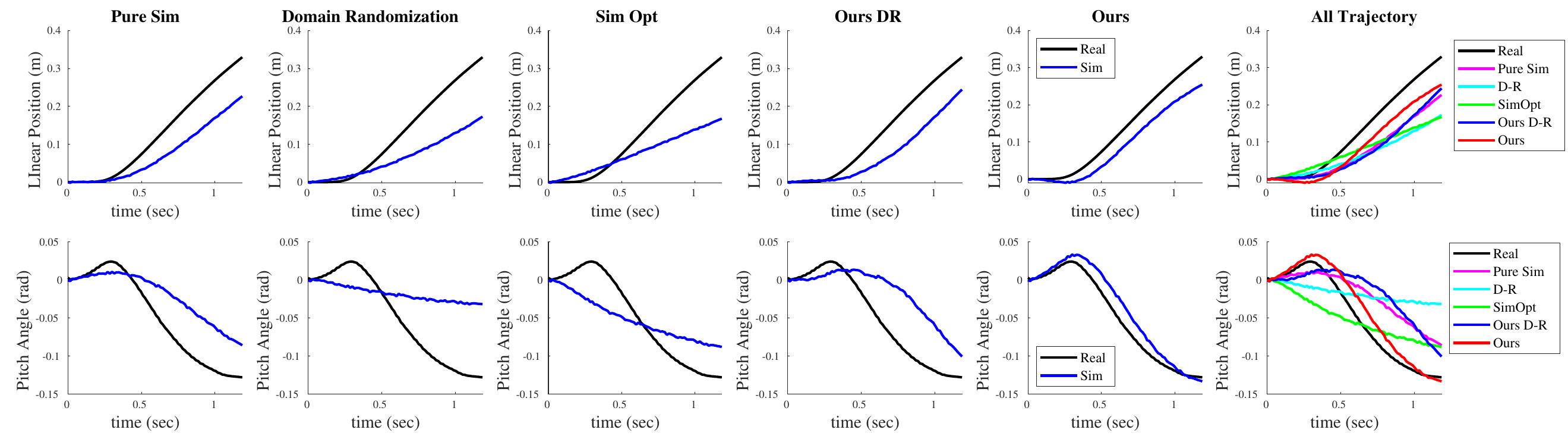}}
    \caption{\textbf{Sample Result of Data Trajectory Comparison Between a Simulation and the Real World.} The last graph presents a comparison of data trajectories between simulation and reality for all methods. Our approach effectively narrows the \textit{reality gap} and aligns the trajectory shape with the actual one. The residual error is attributed from coupled-dynamics, non-parametric nonlinear dynamics, and class-specific data variance. All data samples are collected from the WIP system using LQR control.}
    \label{fig_traj_sample}
\end{figure*}

\subsection{Estimating New Equilibrium Point via Data-Driven Model}
\textbf{Training a Data-Driven Model.} To capture the new equilibrium point for a wheeled-legged robot with unknown dynamics, we focus on the initial behavioral characteristics of a robot under different dynamics influences (e.g., changes in payload weight and center of mass (CoM) position affecting speed during forward falls). This behavior is encapsulated as preconception history, employing time-series features $\bm{X} = [x_{t: N}, \theta_{t: N}]^\top$ as inputs to a data-driven estimator. 

Distinct datasets were constructed from varied simulation configurations, each mirroring the specifics of corresponding baseline real-to-sim methods. A total of 1,500 data points constituted each dataset, divided into training, validation, and testing sets with an 8:1:1 split. The optimum model for each method, validated through simulation datasets, was chosen for benchmarking, as shown in Table \ref{exp1_table_estimation}. The sequence length was established at 80, equivalent to approximately 1.2 seconds of data. The data-driven estimator, exemplified by an LSTM, underwent training exclusively on simulation-derived datasets. Specifically, the LSTM was trained over 500 epochs with a batch size of 256 and a learning rate of $1\times10^{-3}$, alongside a weight decay of $1\times10^{-5}$. The LSTM featured a hidden size of 1024 across two layers.

\textbf{Deployment in a Physical System.}
The pre-trained data-driven model is utilized to predict the new equilibrium point in a physical system without using further manual tuning. The WIP system, when in motion, tends toward the direction where additional mass is added, as illustrated in Fig \ref{testbed}. The estimator quickly identifies the new equilibrium point within a short time frame (less than $1.2$ sec). To facilitate a smooth transition in the desired pitch angle reference, a soft-update method is employed.

\begin{equation}
\begin{split}
    &\theta^{des} = max(\theta_{lin}, -\beta t) \\
    &\theta^{sm}_t = \alpha\theta^{des}_t + (1-\alpha)\theta^{sm}_{t-1}
\end{split}
\label{smooth_update}
\end{equation}
where the first equation encourages that the reference angle updates gradually until the new equilibrium point $\theta_{lin}$. The role of the second equation is to update the reference angle smoothly such as polynomial trajectory. Hyperparameters $\alpha$ and $\beta$ are manually tuned. ($\alpha = 0.05$ and $\beta = 0.1$)
\section{Experiment}
\label{sec:exp_setup}

Three separate experiments are conducted: 1. Verification of real-to-sim adaptation in estimating the new equilibrium point in both a simulation and the real world. 2. Evaluation of the tracking performance of an LQR applied the newly equilibrium point. 3. An ablation study aimed at investigating the influence of the chosen time-series regression model and optimization algorithm.  

\subsection{Simulation Setting}
A RaiSim \cite{hwangbo2018per} simulation is used to collect the simulation dataset $\mathcal{D_S}$. Unified Robot Description Format (URDF) \cite{URDF} file is employed to simulate a customized wheeled inverted pendulum. The control loop runs at a control frequency of approximately 600-700Hz considering the hardware control frequency.

\subsection{Target Prototype Hardware Testbed: Wheeled-Inverted Pendulum}
\begin{figure}[t]
\centering
    \begin{subfigure}[b]{0.19\linewidth}
 		\includegraphics[width=\columnwidth]{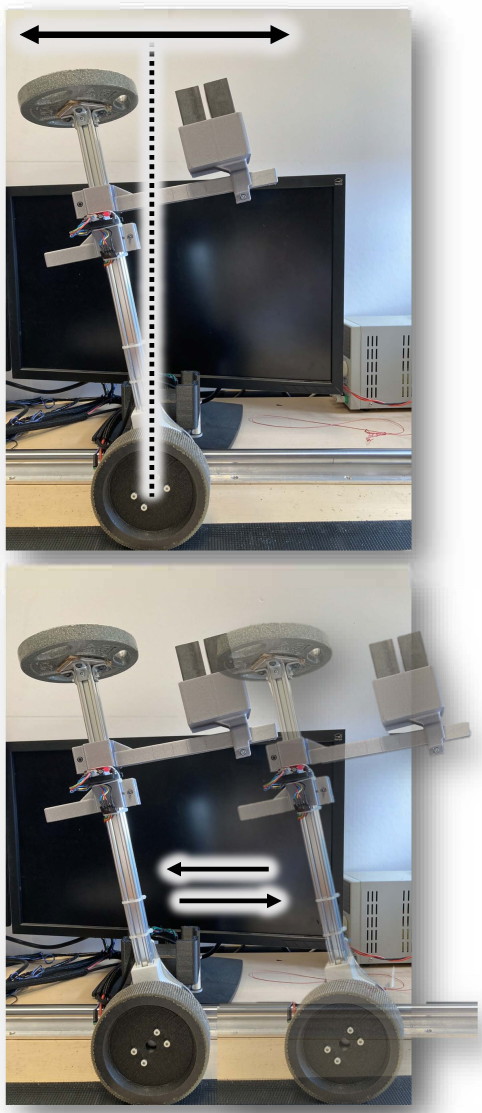}
 		\caption{}
    \end{subfigure} 
    \begin{subfigure}[b]{0.78\linewidth}
 		\includegraphics[width=\columnwidth]{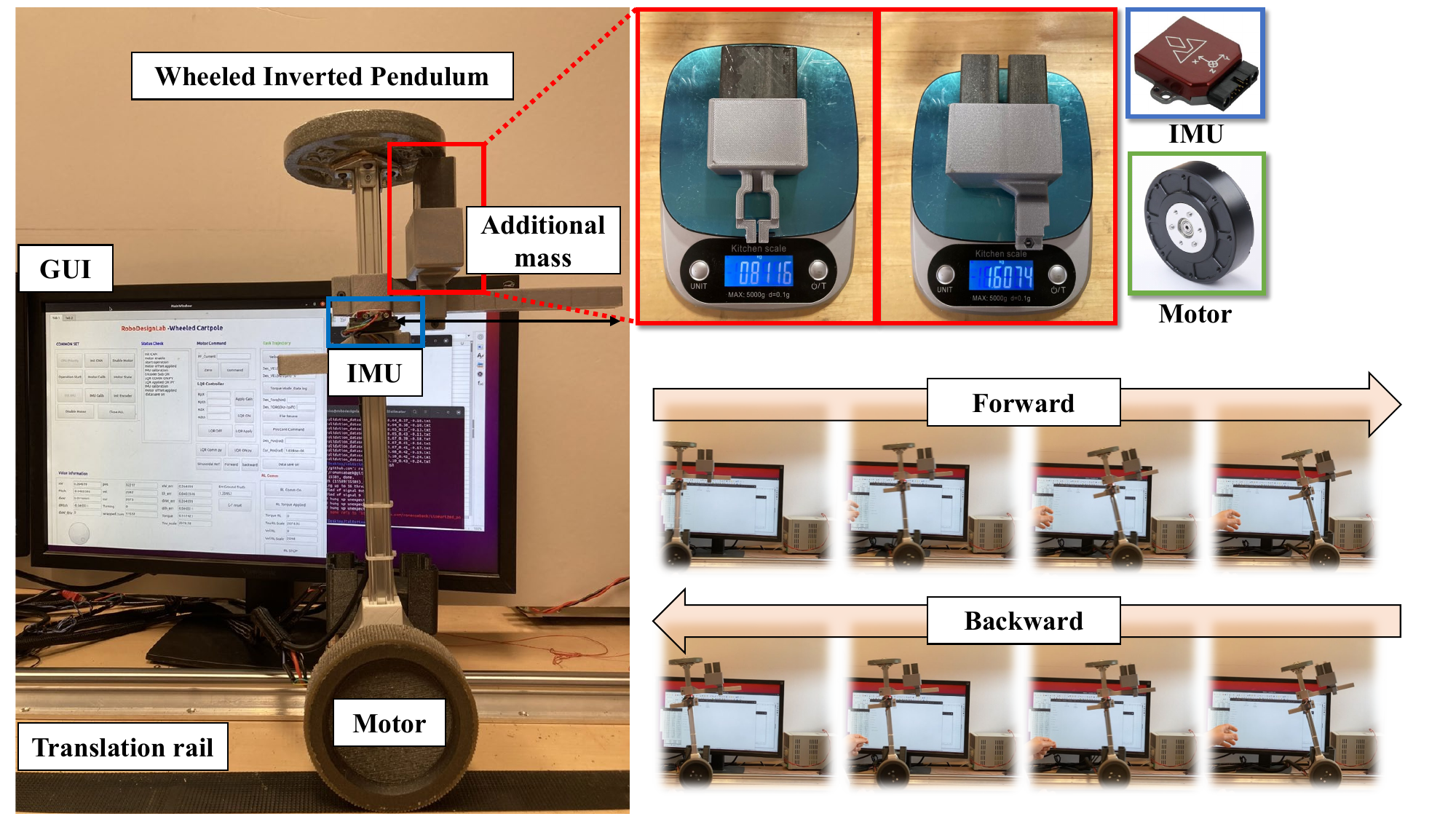}
 		\caption{}
    \end{subfigure} 

    \caption{\textbf{A Physical Wheeled Inverted Pendulum Testbed.} (a) Balancing (top) and tracking a sinusoidal reference task (bottom) were conducted in non-ideal and unknown dynamic situation. (b) A customized wheeled Inverted pendulum is made up of a motor, wheel, IMU, translation rail, and weight. Two additional weights (0.8kg, 1.6kg) can be attached and detached to a pole link to adjust the total mass and the position of the CoM.}
    \label{testbed}
    \vspace{-1.5em}
\end{figure}

A customized wheeled-inverted pendulum (WIP) was developed to verify the feasibility of the proposed framework as depicted in Fig. \ref{testbed}. The WIP is often used as a template model to control a wheeled-legged robot \cite{purushottam2022hands}. Note that while the degree of freedom of WIP is less complex, this still maintains intricate interactions between the robot and its surroundings, including contact dynamics distinct from those of a traditional inverted pendulum. Extra payloads can be affixed to alter its total payload and the position of the CoM. The same actuator equipped in the MIT Mini Cheetah \cite{MiniCheetahICRA} is employed and an inertial measurement unit (VN-100, VectorNav, USA) is mounted to the pole link. The motor communicates over the CAN bus with the desktop PC (Ubuntu20.04) and all software is connected via Robot Operating System (ROS).  

\subsection{Experimental Plan}
\subsubsection{Real-to-Sim Adaptation}
The first experiment was conducted to assess the estimation performance of the data-driven estimator trained in a high-fidelity simulation that is achieved by our proposed framework. The objectives of the experiment are two-fold: (1) to analyze the benefits of a highly precise model in the context of real-to-sim adaptation, and (2) to investigate how our approach stands against traditional domain randomization. We evaluate the impact of the high-fidelity simulation by contrasting it with four separate baselines: (a) \textbf{Pure Sim}: simulation using default physics parameters (b) \textbf{Ours WO Opt}: a more accurate dynamic model (Eq. \ref{nonlinear_f}) with random parameters applied within a simulation (c) \textbf{Sim Param Opt}: built-in simulation parameters adapted through a global optimization algorithm (d) \textbf{Sim Param D-R}: built-in simulation parameters are randomized (\textit{domain randomization}). For the validation, the dataset we described in the section \ref{sec:method1} is utlized.

\subsubsection{Balancing and Tracking Tasks}
The second experiment was conducted to explore the impact of integrating the new equilibrium point into a model-based controller to enhance its tracking performance. The trained data-driven model effectively identified the new equilibrium point online at operation onset by observing the system's behavior under unknown dynamics. This equilibrium point was integrated into an LQR controller, enhancing its ability to track the reference signal and recover the original position of the system. Controller tracking performance was assessed with and without the new equilibrium point in two scenarios: balancing (Task 1) and periodic signal tracking (Task 2), where 'Ours' utilized the new equilibrium point as a target

\begin{table*}[t]
\centering
\setlength{\tabcolsep}{4pt}
\caption{\textbf{Comparison Results of Estimating the Equilibrium Point.} Mean Square Error (MSE) from $146$ test data in simulation (Sim) and $40$ real-world (Real) test data, not involved in training, serves as the evaluation criteria. The discrepancy in MSE between a simulation (Sim) and real world (Real) assesses the model's capability to adapt simulation-trained data to real-world scenarios.}
\begin{tabular}{cc|cccccccccc}
\hline
\multicolumn{2}{c|}{}                            & \multicolumn{2}{c|}{Pure Sim} & \multicolumn{2}{c|}{Ours WO Opt} & \multicolumn{2}{c|}{Sim Param Opt} & \multicolumn{2}{c|}{Sim Param D-R} & \multicolumn{2}{c}{\textbf{Ours}}  \\ \hline
\multicolumn{2}{c|}{}                            & Sim           & Real          & Sim             & Real           & Sim             & Real             & Sim                  & Real        & Sim          & Real                \\ \hline
\multicolumn{1}{c|}{\multirow{2}{*}{MSE (\textit{rad})}} & mean & 0.012         & 0.033         & 0.011           & 0.030          &0.089                  &0.096                   & \textbf{0.008}       & 0.111       & 0.024        & \textbf{0.023}      \\
\multicolumn{1}{c|}{}                     & std  & 0.010         & 0.003         & 0.007           & 0.005          &0.063                 &0.025                 & 0.008                & 0.010       & 0.015        & 0.005               \\ \hline
\multicolumn{2}{c|}{Difference}                  & \multicolumn{2}{c}{0.021}     & \multicolumn{2}{c}{0.019}        & \multicolumn{2}{c}{0.007}               & \multicolumn{2}{c}{0.103}          & \multicolumn{2}{c}{\textbf{0.001}} \\ \hline
\end{tabular}
\label{exp1_table_estimation}
\end{table*}

\begin{figure*}[t]
    \centering
    \begin{subfigure}[b]{0.195\linewidth}
 		\includegraphics[width=\columnwidth]{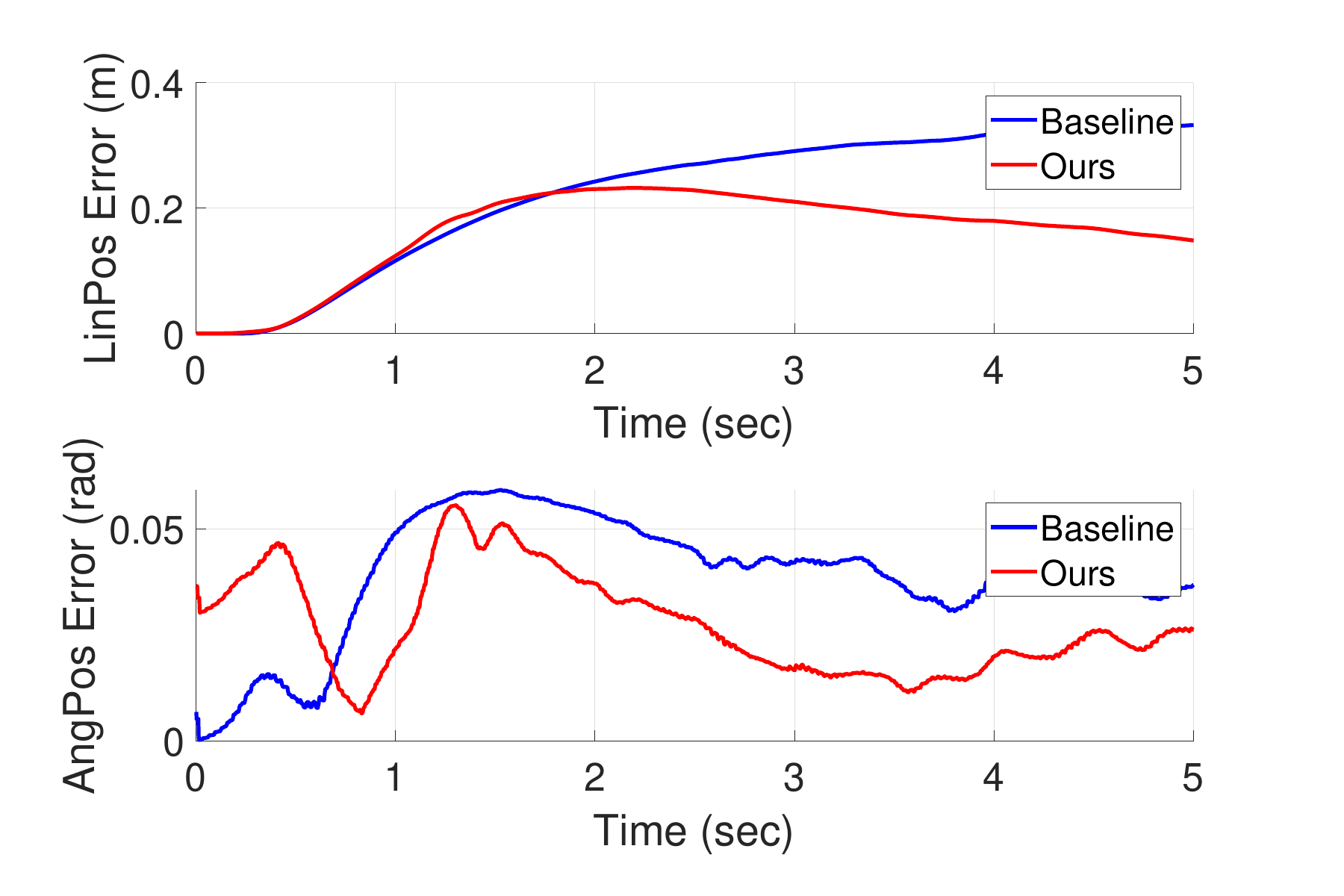}
 		\caption{$\theta_{lin} = -0.03rad$}
    \end{subfigure} 
    \begin{subfigure}[b]{0.195\linewidth}
 		\includegraphics[width=\columnwidth]{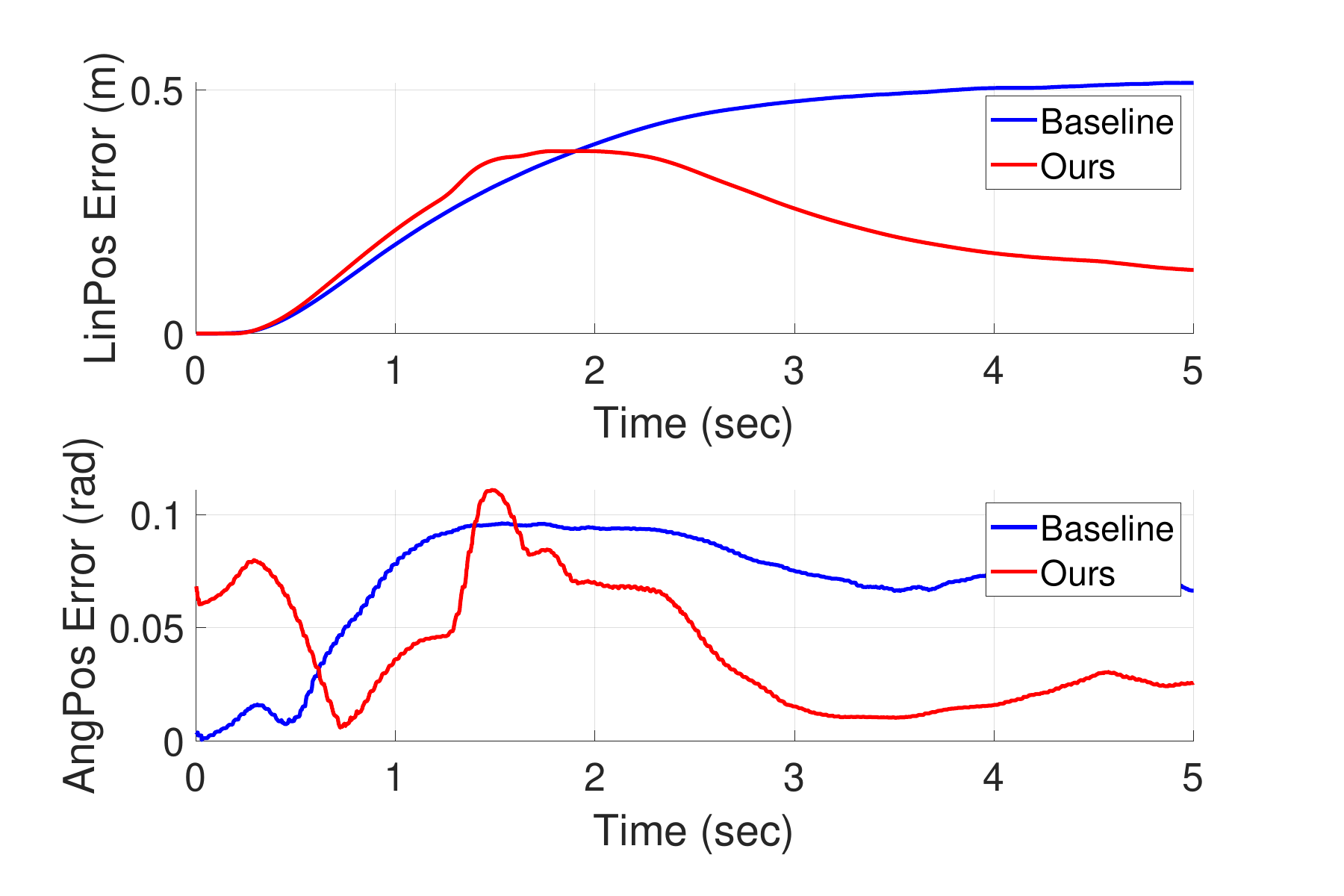}
 		\caption{$\theta_{lin} = -0.06rad$}
 	\end{subfigure} 
    \begin{subfigure}[b]{0.195\linewidth}
 		\includegraphics[width=\columnwidth]{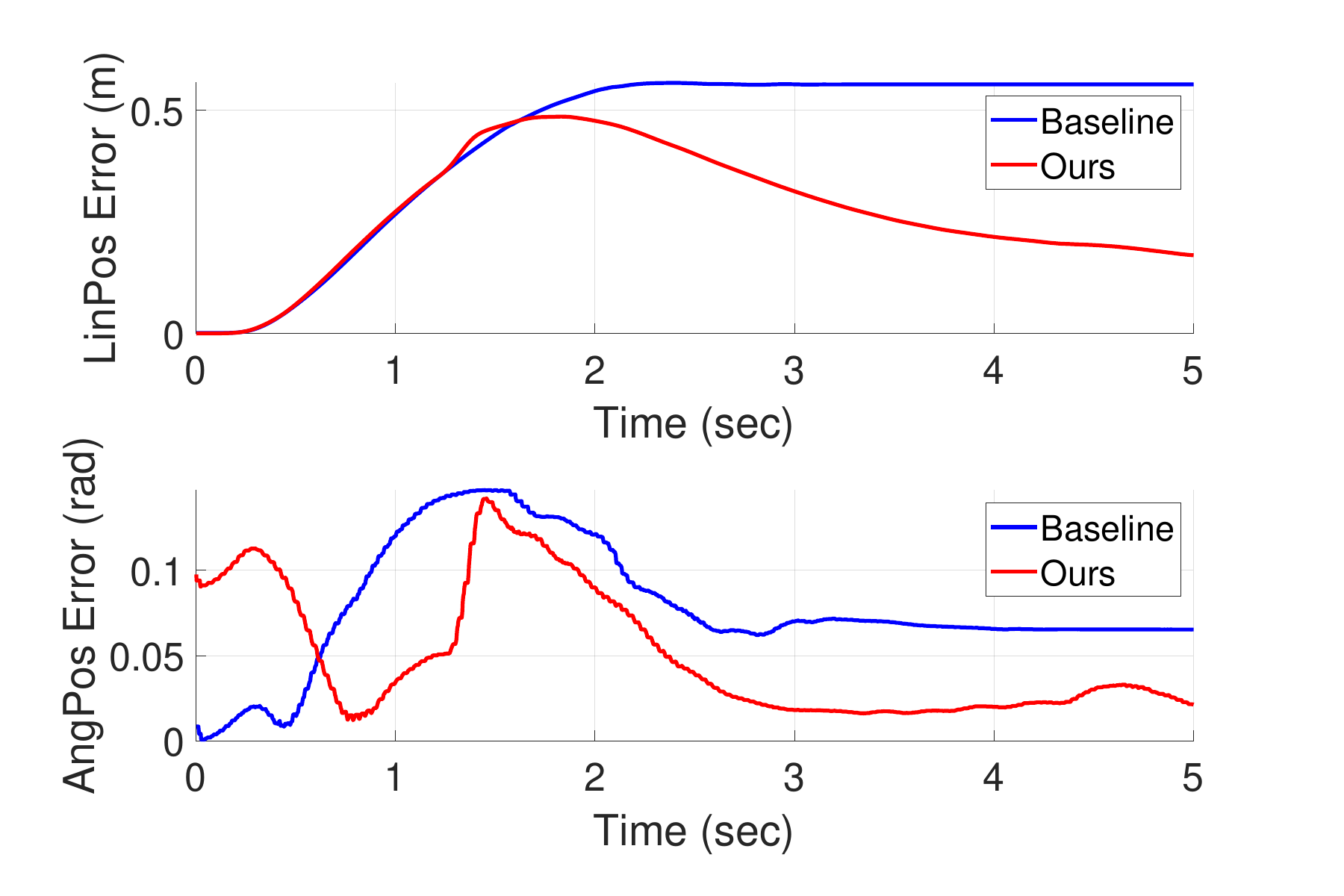}
 		\caption{$\theta_{lin} = -0.09rad$}
    \end{subfigure} 
    \begin{subfigure}[b]{0.195\linewidth}
 		\includegraphics[width=\columnwidth]{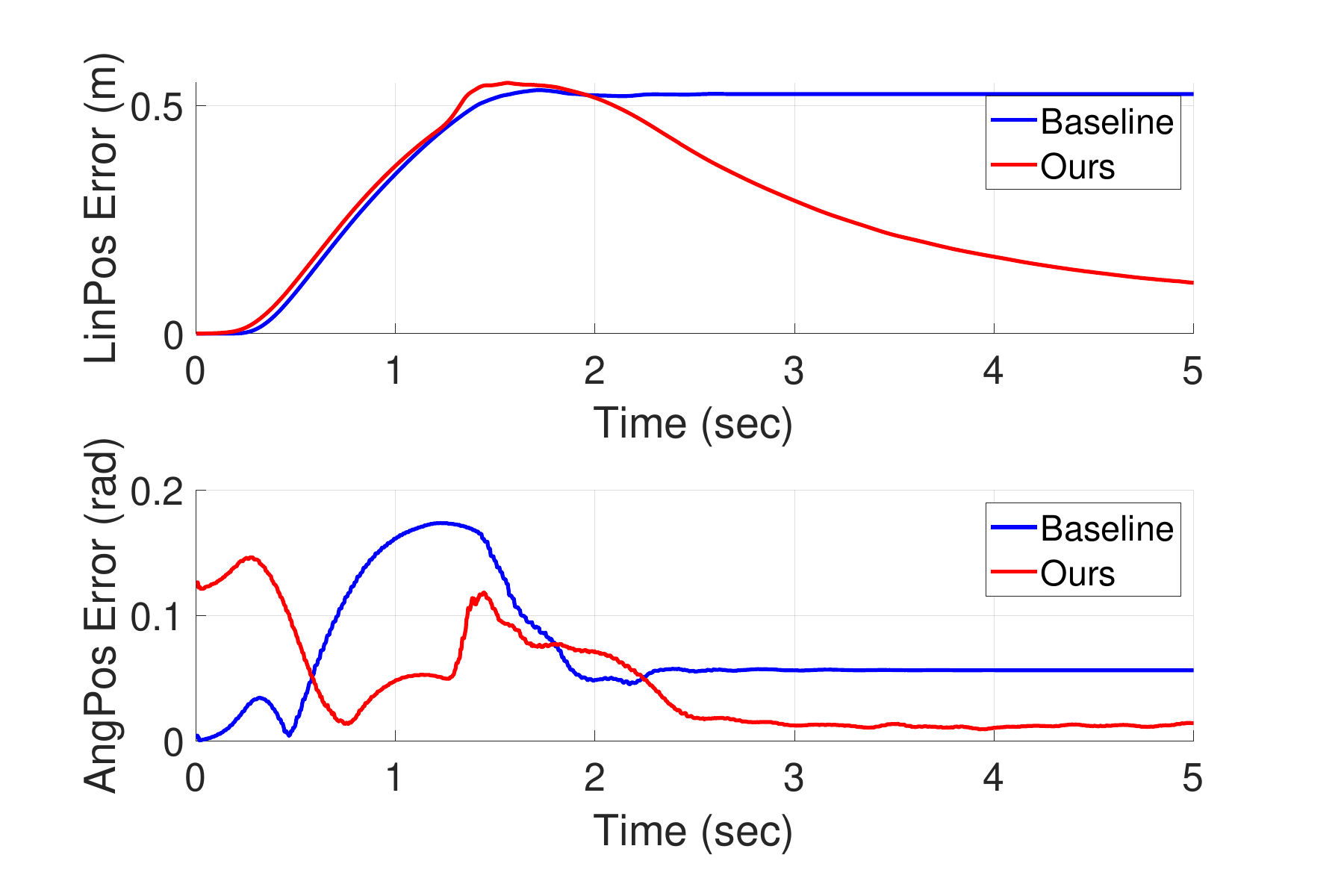}
 		\caption{$\theta_{lin} = -0.12rad$}
    \end{subfigure} 
    \begin{subfigure}[b]{0.195\linewidth}
 		\includegraphics[width=\columnwidth]{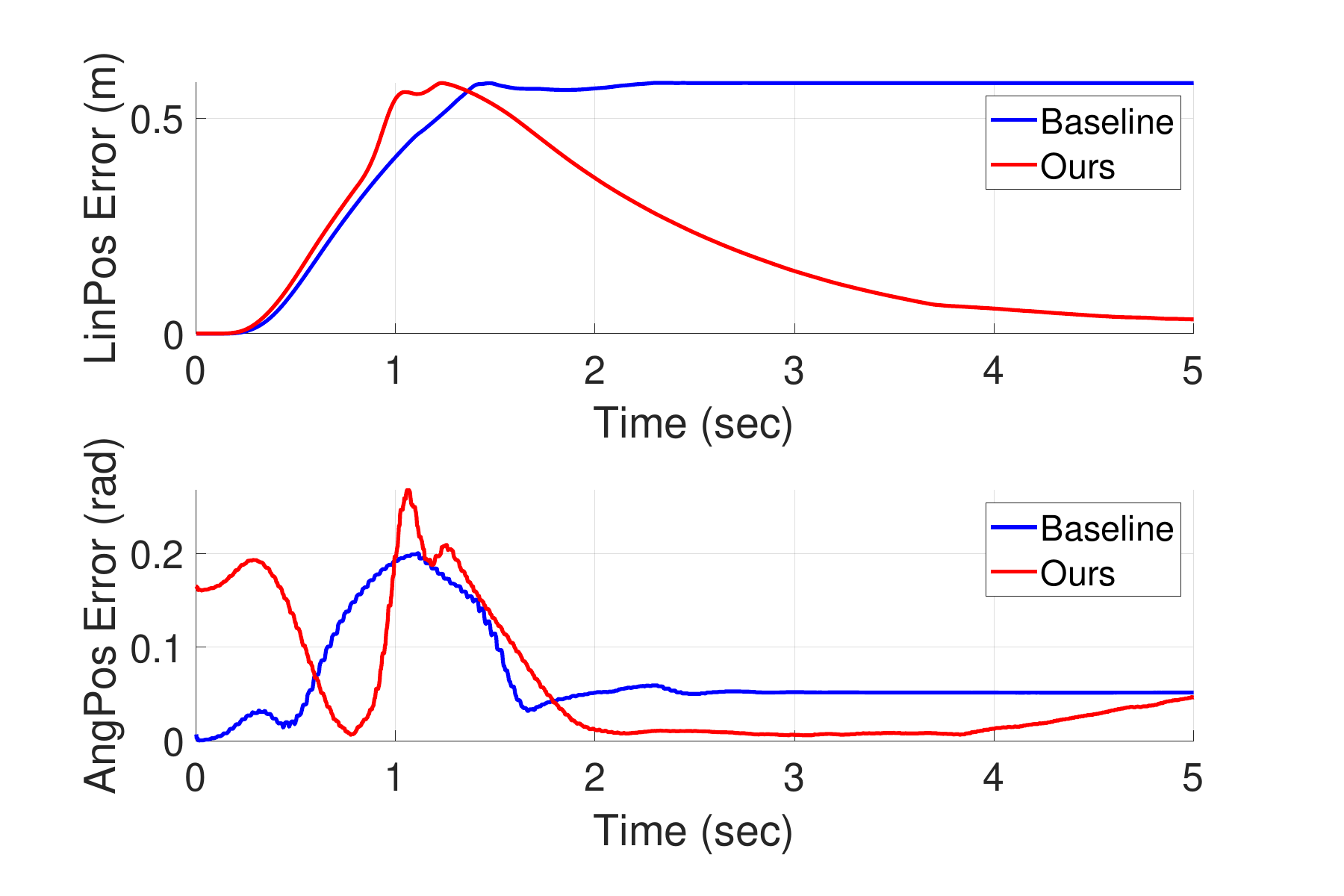}
 		\caption{$\theta_{lin} = -0.16rad$}
    \end{subfigure} 
    \caption{\textbf{Results for Balancing Tasks in Several Scenarios with Different, Unknown Dynamics.} Graphs (a) to (e) display the absolute error between targeted and actual trajectories for both linear ($x_w$) and angular ($\theta$) positions (mean of five trails). The baseline assumes zero for both desired positions, while our method employs the new equilibrium point as the targeted angular position. Each graph corresponds to a unique equilibrium point, reflecting varied payloads and the center of mass. The baseline LQR tends to fail, struggling with equilibrium points significantly deviated from the original (When
    the system reached and crashed the end of rail, the system halt at a certain point - this is why the baseline graph appears to stop at a certain point). We observed the system continuously fell until reaching its motion limit. Conversely, our method quickly identified the new equilibrium point (within approximately 1 second) and effectively recovered its position in all different cases.}
    \label{exp2_task1_result}
\end{figure*}

\begin{figure*}[t]
    \centering
   \begin{subfigure}[b]{0.195\linewidth}
 		\includegraphics[width=\columnwidth]{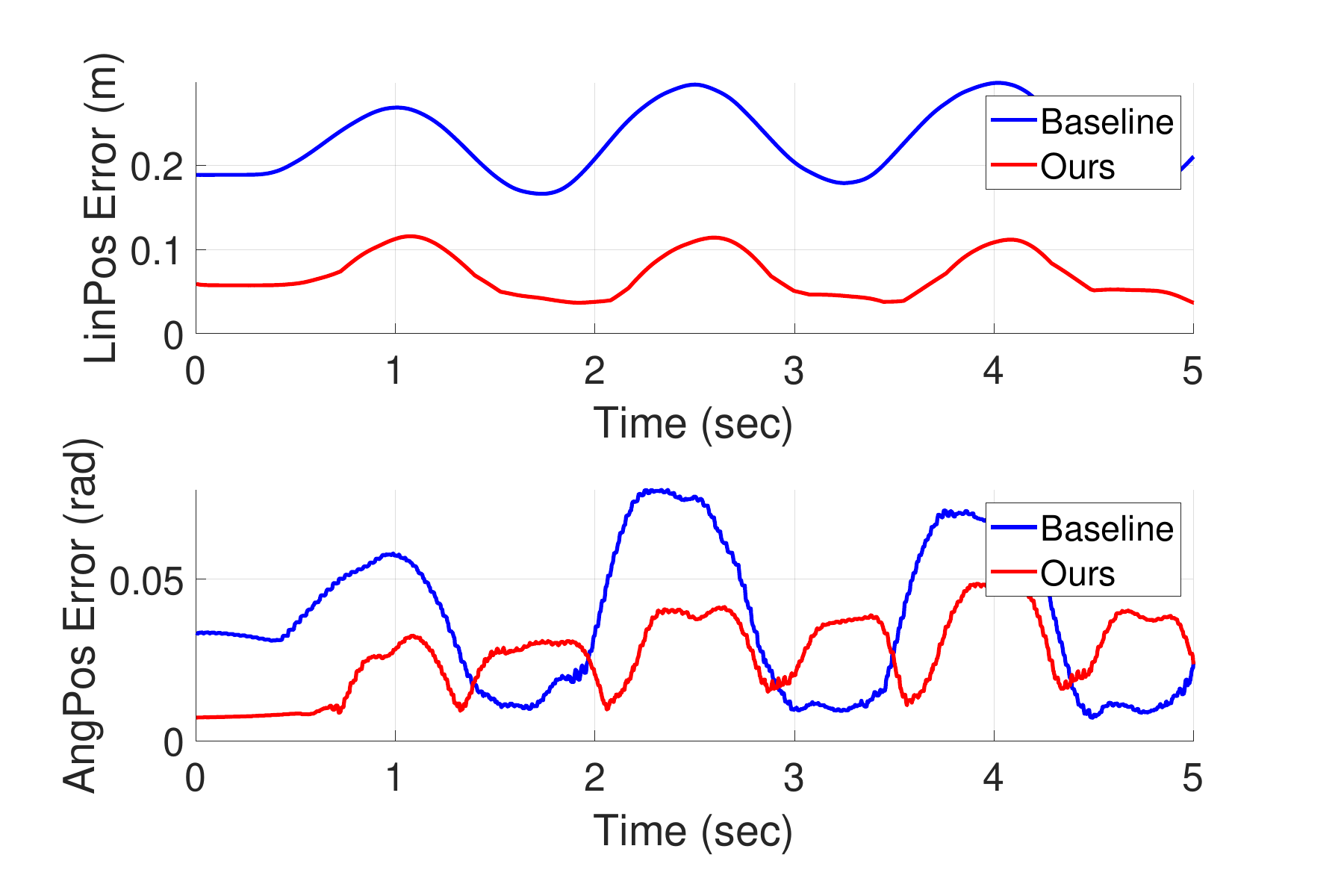}
 		\caption{$\theta_{lin} = -0.03rad$}
    \end{subfigure} 
    \begin{subfigure}[b]{0.195\linewidth}
 		\includegraphics[width=\columnwidth]{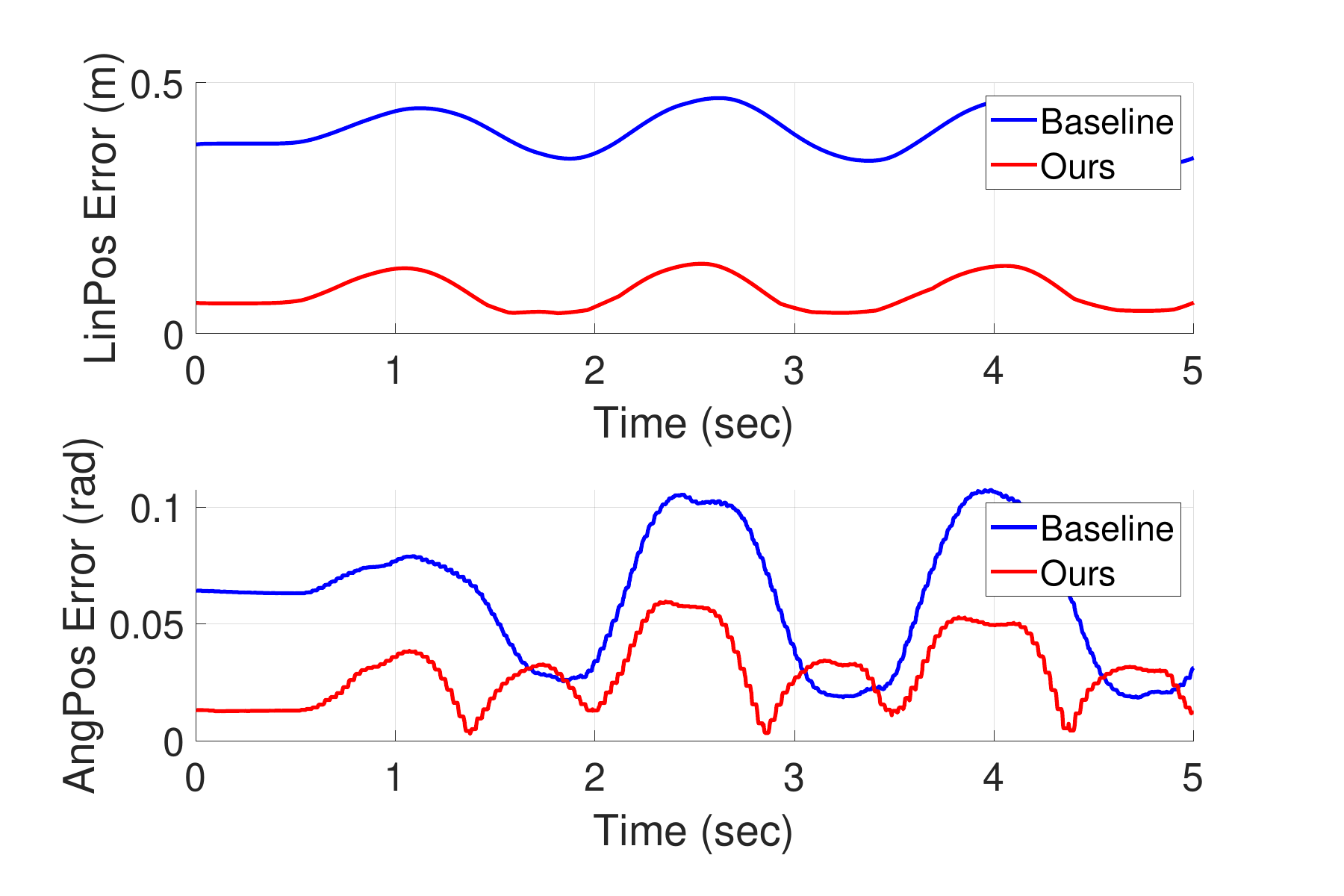}
 		\caption{$\theta_{lin} = -0.06rad$}
 	\end{subfigure} 
    \begin{subfigure}[b]{0.195\linewidth}
 		\includegraphics[width=\columnwidth]{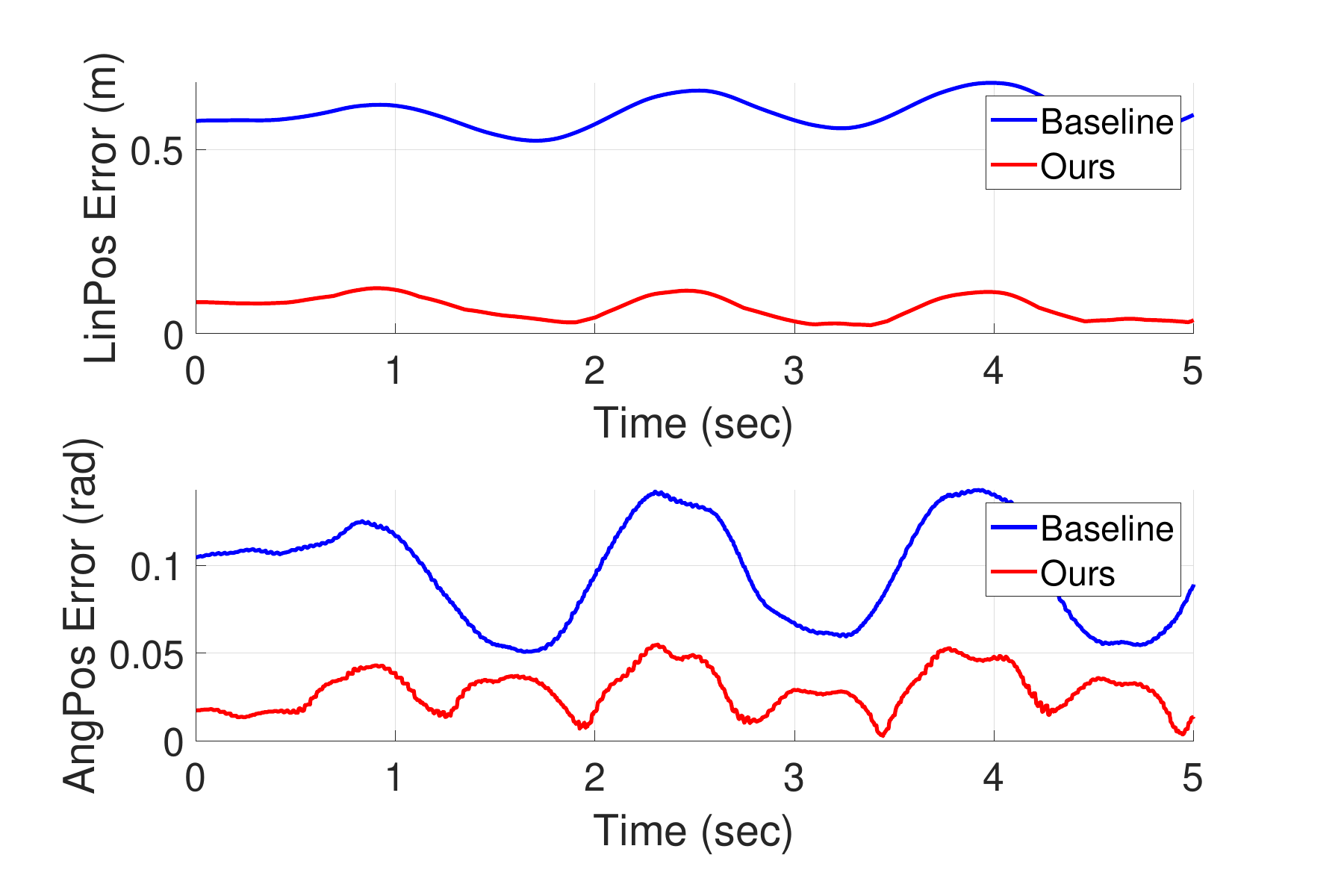}
 		\caption{$\theta_{lin} = -0.09rad$}
    \end{subfigure} 
    \begin{subfigure}[b]{0.195\linewidth}
 		\includegraphics[width=\columnwidth]{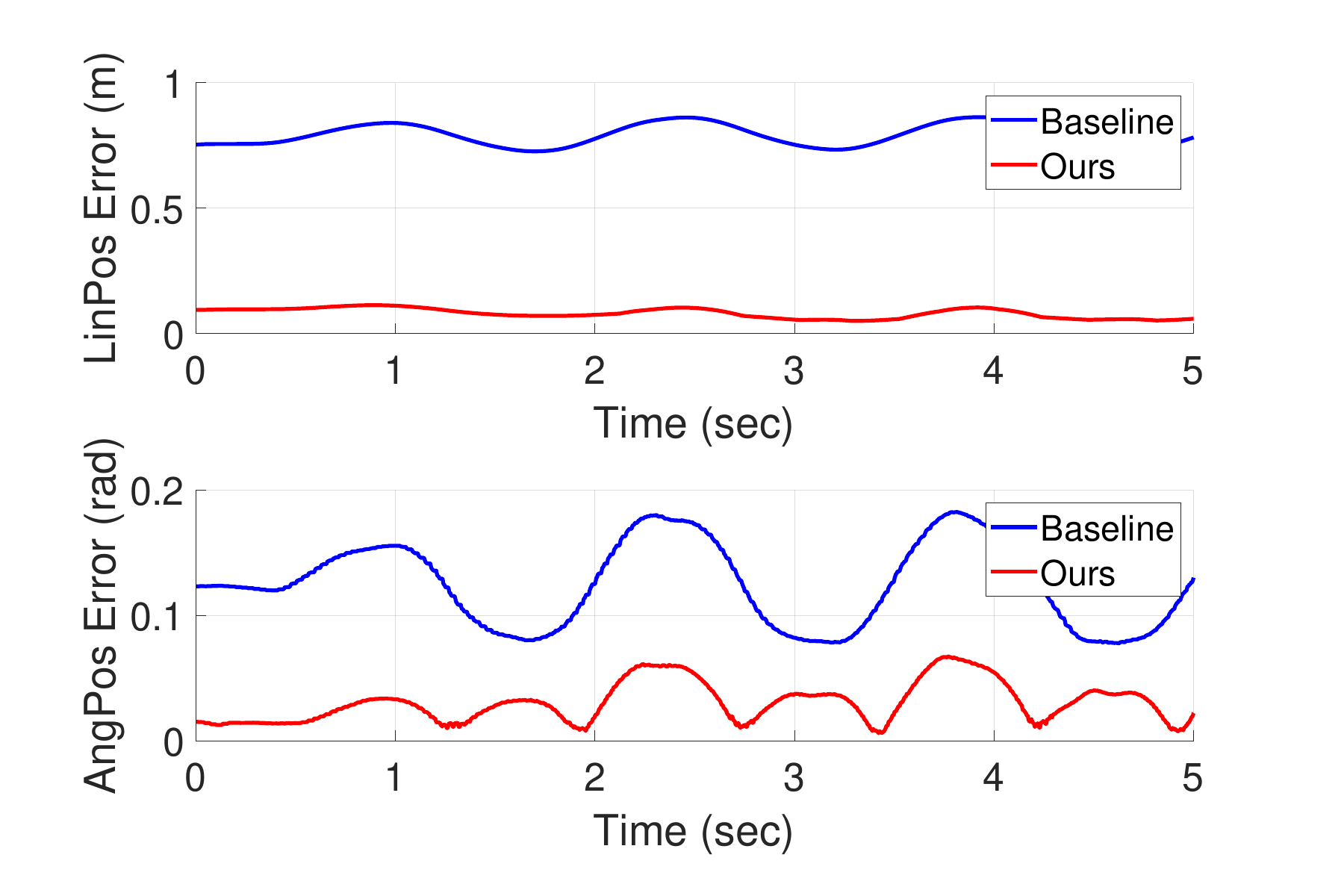}
 		\caption{$\theta_{lin} = -0.12rad$}
    \end{subfigure} 
    \begin{subfigure}[b]{0.195\linewidth}
 		\includegraphics[width=\columnwidth]{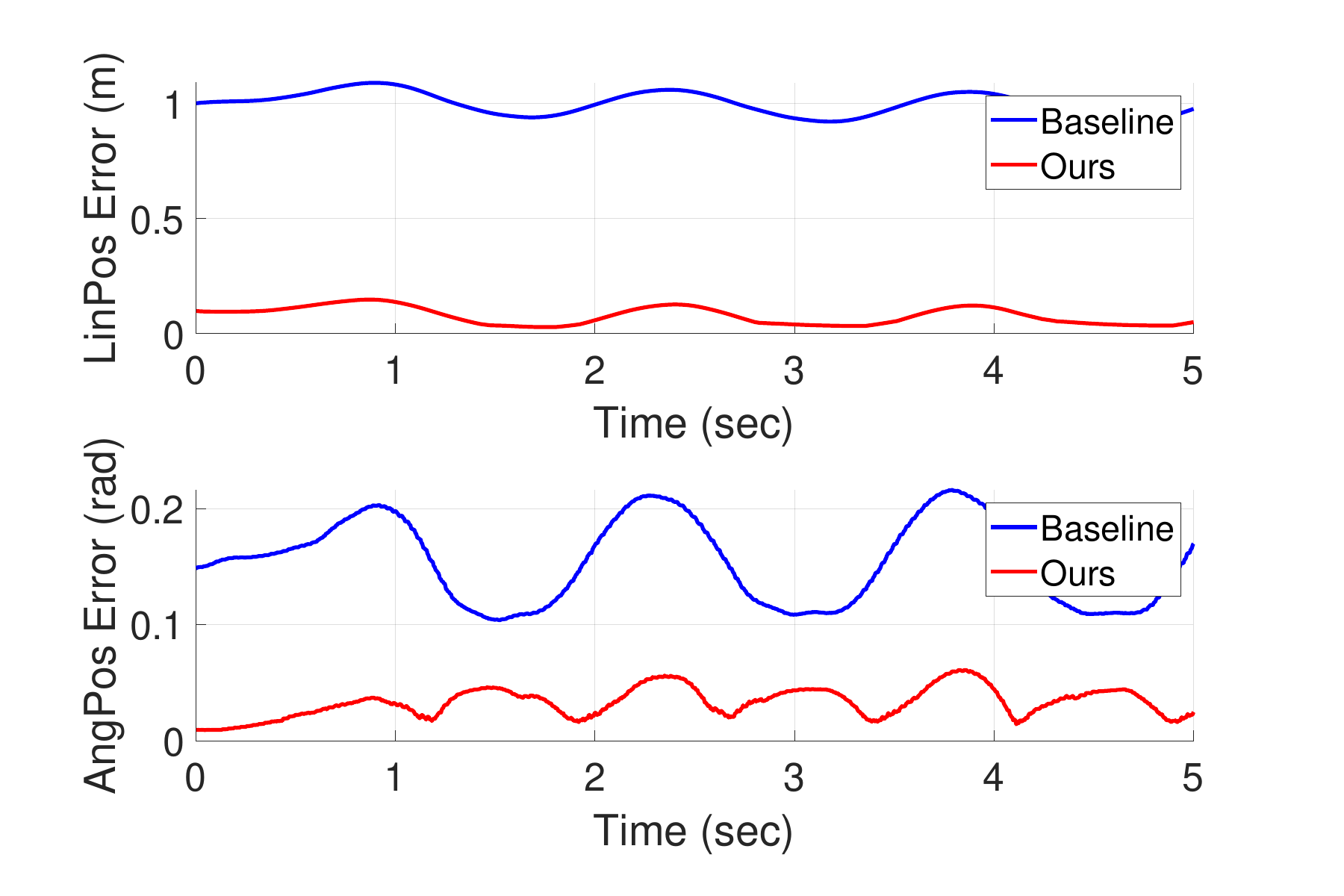}
 		\caption{$\theta_{lin} = -0.16rad$}
    \end{subfigure} 

    \caption{\textbf{Results for Tracking Tasks in Several Scenarios with Different, Unknown Dynamics.} Graphs (a) to (e) show the same criteria we mentioned in Fig. \ref{exp2_task1_result} for the tracking task. The periodic (sinusoidal) signal is utilized as a desired reference. Incorporating a new equilibrium point as the desired angular position, our method demonstrated superior tracking accuracy for both linear and angular positions across all cases.} 
    \label{exp2_task2_result}
\end{figure*}

\noindent angular position, and 'Baseline' did not. We adjusted the total mass and CoM position of the WIP by adding weights to the pole link, simulating a wheeled-legged robot transporting an unknown object. Two task experiments were conducted across five different cases, with each case undergoing five trials. 
\section{Result and Discussion}
\label{result}

\subsection{Experiment 1: Estimating New Equilibrium Point in WIP with Unknown Dynamics}

As shown in Fig. \ref{fig_traj_sample}, using a more accurate nonlinear dynamics model \ref{nonlinear_f} showed its benefit in narrowing the \textit{reality gap} in terms of position trajectories. Notably, on the graph below (pitch angle), only our method showed a similar pattern to the target graph (initially spiking upwards). Although two trajectories is not perfectly overlapped in ours case, results from our following experiments support that this is sufficient to enhance the performance of estimating the new equilibrium point (MSE less than $1.5 rad$). The residual error might be caused by the other parametric errors (e.g., inertia of each link), coupled-dynamics, and non-parametric modeling error such as backlash and hysteresis.

Validation outcomes for the new equilibrium point estimation are summarized in Table \ref{exp1_table_estimation}. These results demonstrate the efficacy of the data-driven estimator (e.g., LSTM), trained within a high-fidelity simulation from our framework, in narrowing the \textit{reality gap} and thus improving new equilibrium point estimation accuracy. This underscores the importance of incorporating an elaborate dynamics model and optimizing its parameters for a seamless sim-to-real dynamics transfer. Notably, the MSE discrepancy between simulation (Sim) and real-world (Real) cases is a mere $0.001$ rad.

\textbf{Sim Param D-R} counters \textit{over-fitting}, improving estimation in simulation tests but performing poorly in real scenarios, as \textit{Domain Randomization} doesn't directly bridge the model trajectory mismatch \cite{li2023robust}. Generally, this can enhance model robustness by reducing sensitivity to non-essential features like noise, however, this attributes to increase dataset variance through parameter diversification, bringing the performance degradation in the regression problem. \textbf{Ours WO Opt} shows marginally better estimation over \textbf{Pure Sim}, suggesting the integration of a refined dynamics model (Eq. (\ref{nonlinear_f})) helps narrow the \textit{reality gap}, despite randomized parameters in training. \textbf{Sim Param Opt} shows the least accuracy, indicating that optimizing simulation parameters barely reduces the cost for 
$\br{\zeta}$ and hampers extracting vital features for new equilibrium point estimation, thus degrading estimation accuracy.

\subsection{Experiment 2: Control Performance Validation} 
\subsubsection{Balancing Task with Unknown Dynamics}
In the balancing task with an unknown payload, utilizing a new equilibrium point effectively recover the WIP system's original position, as depicted in Fig. \ref{exp2_task1_result}. More importantly, we observed that the WIP using the baseline LQR was prone to  toppling over and continued to advance until reaching the mechanical limit of the rail. This shows that our method goes beyond simply increasing the tracking performance and contributes to improving the stability of the system. On the average considering all five cases, we observed 38\% and 23\% performance improvement in root mean square error of a position and angular velocity, respectively. The performance improvement in each case is as follows: [24.34\%, 40.87\%, 35.29\%, 52.60\%, 31.24\%, 47.51\%] in a position and  [0.2574\%, 0.3377\%, 0.2837\%, 0.3542\%, 0.2759\%, -0.1076\%] in an angular position.

\subsubsection{Tracking Task with Unknown Dynamics}
In the tracking task, we used a predefined sinusoidal signal as a desired velocity ($\dot{xW}_{des}=0.3sin(2\pi ft), f=0.4$). Given that the baseline LQR fails at the operation's onset with an unknown payload attached to the system (see Fig. \ref{exp2_task1_result}), we began the tracking task once the system stabilizes over time for a fair tracking task experiment. As shown in Fig. \ref{exp2_task2_result}, using a new equilibrium point from our estimator led to the tracking performance improvement in both a position and angular position. On average, across all five cases, there was a 77\ improvement in the root mean square error for position and a 60\% improvement for angular velocity.  The performance improvement in each case is as follows: [0.6646\%, 0.7493\%, 0.8588\%, 0.5962\%, 0.8811\%, 0.9065\%] in a position and [0.3407\%, 0.5073\%, 0.6740\%, 0.6051\%, 0.7142\%, 0.7595\%] in an angular position.

\subsection{Ablation Study}
\subsubsection{Performance Comparison of Different Optimization Algorithm.}
\begin{figure}[t]
    \centering
 	\begin{subfigure}[b]{0.48\linewidth}
 		\includegraphics[width=\columnwidth]{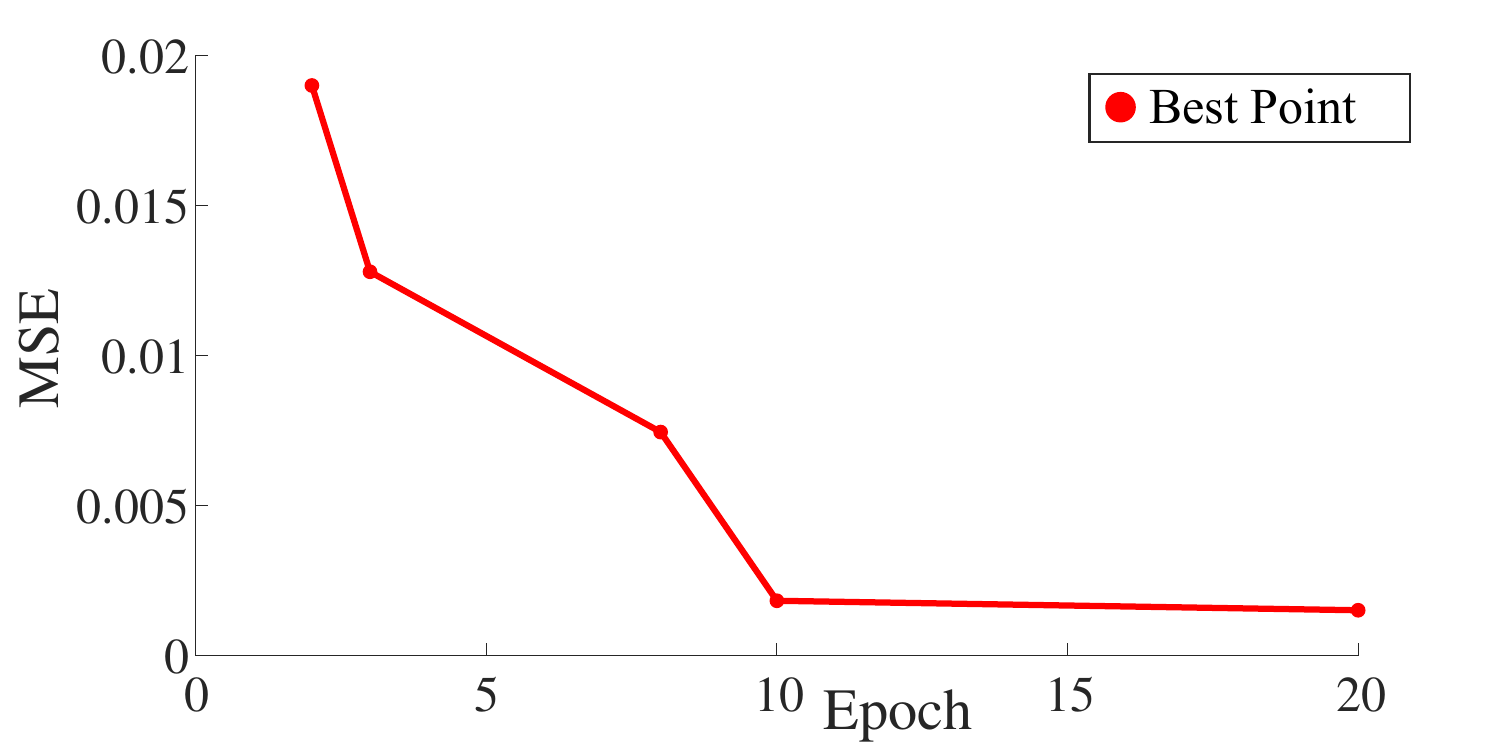}
 		\caption{GA}
 	\end{subfigure} 
    \begin{subfigure}[b]{0.48\linewidth}
 		\includegraphics[width=\columnwidth]{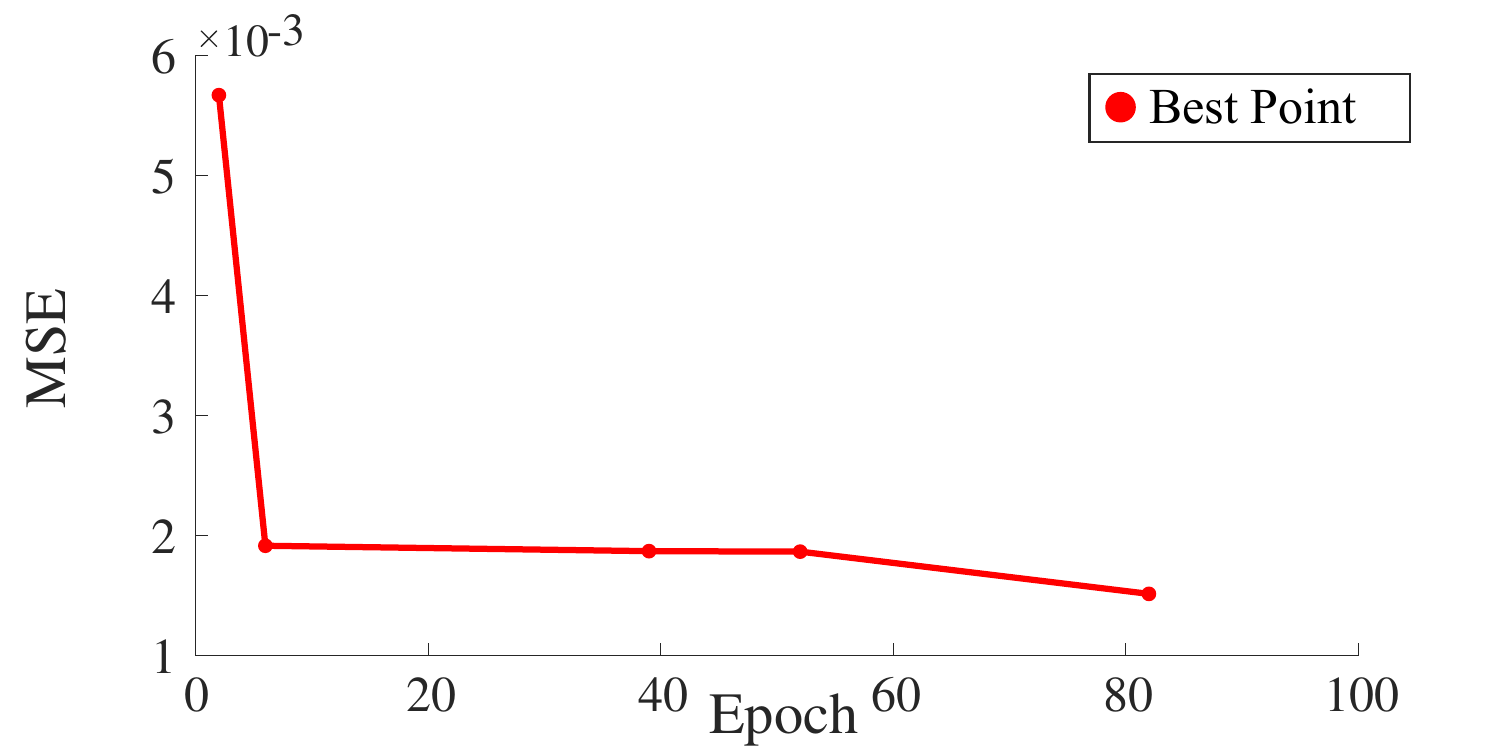}
 		\caption{PSO}
 	\end{subfigure} 
\caption{\textbf{Result of MSE according to update the parameter}. Graphs (a) and (b) compare the MSE (Eq. (\ref{costfunc})) results of the GA and PSO, respectively, showing PSO's faster update rate allows it to achieve more epochs within the same timeframe.}
\label{append_optim}
\end{figure}

An additional experiment assessed the best optimization algorithm for identifying the nonlinear model parameters 
$\br{\zeta}$, focusing on accuracy and time efficiency. Among seven algorithms tested using the SciPy package \cite{virtanen2020scipy}—genetic algorithm (GA), particle swarm optimization (PSO), Nelder–Mead (NM), Powell, conjugate gradient (CG), BFGS, and sequential least squares programming (SLSQP)—only GA and PSO succeeded in parameter optimization, as shown in Fig. \ref{append_optim}. This success likely stems from their global optimization capabilities, crucial for addressing the non-convex nature of our problem, whereas the others may stall at local minima. PSO outperformed GA in both convergence speed and accuracy, benefiting from its simpler velocity and position update processes. Optimizations were halted if exceeding 30 minutes, using default hyperparameters for each method.

\subsubsection{Performance Benchmark of Time-Series Data-Driven Model.}
In Table \ref{apped_table}, we evaluate the new equilibrium point estimation performance using various time-series data-driven methods, also assessing the effect of sequence data length on accuracy. These methods were trained with high-fidelity simulations from our framework. Our results indicate improved performance with more historical data, suggesting initial WIP system movements provide insufficient features for identifying the new equilibrium point across different weights and CoM positions. TCN achieved the highest accuracy and fastest inference with shorter history data lengths. Conversely, LSTM excelled with 80 data points, potentially due to TCN's concurrent processing advantage and its structural benefits in addressing gradient issues common in recurrent networks. While 1D-CNN showed superior simulation data results, its real-world performance was less impressive. Notably, the choice of method did not critically affect our application's outcome.

\subsection{Limitation}
Although our method is more sample-efficient than model-free RL, it still requires a target domain dataset for Real-to-Sim adaptation. Collecting this dataset is challenging for high-dimensional systems with unknown payloads. Developing a Real-to-Sim adaptation method that does not rely on a target dataset is an interesting future direction. Additionally, to enhance simulation fidelity, errors from non-parametric models should be addressed using techniques like Gaussian Processes and Deep Neural Networks, which are commonly employed to represent such models.

\begin{table}[t]
\centering
\setlength{\tabcolsep}{1pt}
\caption{\textbf{Performance benchmark for estimating the new equilibrium point between five different data-driven model.} Mean Square Error (MSE) of $146$ test data obtained from a simulation (Sim) and $40$ data obtained from the real world (Real). Length indicates the length of data trajectories which are an input of the data-driven estimator.}
\begin{tabular}{cc|clcl|clcl|clll|clll|clcl}
\hline
\multicolumn{2}{c|}{}                                                               & \multicolumn{4}{c|}{\textbf{LSTM}}                              & \multicolumn{4}{c|}{\textbf{Transformer}}              & \multicolumn{4}{c|}{\textbf{GRU}}                      & \multicolumn{4}{c|}{\textbf{TCN}}                               & \multicolumn{4}{c}{\textbf{1D-CNN}}                                      \\ \hline
\multicolumn{2}{c|}{}                                                               & \multicolumn{2}{c}{Sim}   & \multicolumn{2}{c|}{Real}           & \multicolumn{2}{c}{Sim}   & \multicolumn{2}{c|}{Real}  & \multicolumn{2}{c}{Sim}   & \multicolumn{2}{c|}{Real}  & \multicolumn{2}{c}{Sim}   & \multicolumn{2}{c|}{Real}           & \multicolumn{2}{c}{Sim}            & \multicolumn{2}{c}{Real}           \\ \hline
\multicolumn{1}{c|}{\multirow{4}{*}{Length}}                  & 20                  & \multicolumn{2}{c}{0.038} & \multicolumn{2}{c|}{\textbf{0.036}} & \multicolumn{2}{c}{0.064} & \multicolumn{2}{c|}{0.060} & \multicolumn{2}{c}{0.010} & \multicolumn{2}{c|}{0.037} & \multicolumn{2}{c}{0.068} & \multicolumn{2}{c|}{0.065}          & \multicolumn{2}{c}{\textbf{0.009}} & \multicolumn{2}{c}{0.040}          \\
\multicolumn{1}{c|}{}                                         & 40                  & \multicolumn{2}{c}{0.031} & \multicolumn{2}{c|}{0.042}          & \multicolumn{2}{c}{0.064} & \multicolumn{2}{c|}{0.060} & \multicolumn{2}{c}{0.008} & \multicolumn{2}{l|}{0.036} & \multicolumn{2}{l}{0.087} & \multicolumn{2}{l|}{0.065}          & \multicolumn{2}{l}{\textbf{0.005}} & \multicolumn{2}{c}{\textbf{0.033}} \\
\multicolumn{1}{c|}{}                                         & 60                  & \multicolumn{2}{c}{0.012} & \multicolumn{2}{c|}{0.030}          & \multicolumn{2}{c}{0.063} & \multicolumn{2}{c|}{0.060} & \multicolumn{2}{c}{0.010} & \multicolumn{2}{l|}{0.032} & \multicolumn{2}{l}{0.020} & \multicolumn{2}{l|}{\textbf{0.023}} & \multicolumn{2}{l}{\textbf{0.005}} & \multicolumn{2}{c}{0.033}          \\
\multicolumn{1}{c|}{}                                         & 80                  & \multicolumn{2}{c}{0.024} & \multicolumn{2}{c|}{\textbf{0.023}} & \multicolumn{2}{c}{0.028} & \multicolumn{2}{c|}{0.024} & \multicolumn{2}{c}{0.020} & \multicolumn{2}{l|}{0.030} & \multicolumn{2}{l}{0.038} & \multicolumn{2}{l|}{0.027}          & \multicolumn{2}{l}{\textbf{0.005}} & \multicolumn{2}{c}{0.035}          \\ \hline
\multicolumn{2}{c|}{\begin{tabular}[c]{@{}c@{}}Inference \\ Time (sec)\end{tabular}} & \multicolumn{4}{c|}{0.035}                                      & \multicolumn{4}{c|}{\textbf{0.003}}                    & \multicolumn{4}{c|}{0.02}                              & \multicolumn{4}{c|}{\textbf{0.003}}                             & \multicolumn{4}{c}{0.006}                                               \\ \hline
\end{tabular}
\label{apped_table}
\end{table}

\section{Conclusion}
\label{concl}

In this work, we present a framework designed to control a wheeled humanoid robot with unknown dynamics, aimed at safely and accurately delivering unknown objects. Ultimately, to facilitate system stability analysis and improve model accuracy, our approach is to estimate the new equilibrium of the system explicitly and utilize it in a model-based controller. Data-driven method is utilized to rapidly estimate the equilibrium point of the system with unknown dynamics. To be more efficient in the data collection process, we propose a novel real-to-sim adaptation that is capable of reducing the \textit{reality gap} at the beginning of training a data-driven model and constructed dataset in the high-fidelity simulation. We conducted experiments with a physical inverted pendulum we developed as a simplified version of a wheeled humanoid. The results suggested that using a more accurate nonlinear dynamics with its optimized parameter showed benefit in narrowing the \textit{reality gap}, contributing to improving the tracking performance of a model-based controller. 



\bibliographystyle{unsrt}
\bibliography{main.bib}

\end{document}